\documentclass{article} 
\usepackage{iclr2024_conference,times}

\usepackage{amsmath,amsfonts,bm}

\def\eqref#1{equation~\ref{#1}}

\def\1{\bm{1}}

\DeclareMathAlphabet{\mathsfit}{\encodingdefault}{\sfdefault}{m}{sl}
\SetMathAlphabet{\mathsfit}{bold}{\encodingdefault}{\sfdefault}{bx}{n}

\iclrfinalcopy

\usepackage{hyperref}
\usepackage{url}
\usepackage{natbib}
\usepackage{booktabs}
\usepackage{graphicx}
\usepackage{comment}
\usepackage{amsmath}

\title{Improving Interpersonal Communication by Simulating Audiences with Language Models}

\author{Ryan Liu\textsuperscript{\textnormal{1}}, ~Howard Yen\textsuperscript{\textnormal{1}}, ~Raja Marjieh\textsuperscript{\textnormal{2}}, ~Thomas L. Griffiths\textsuperscript{\textnormal{1,2}}, ~Ranjay Krishna\textsuperscript{\textnormal{3}}\\
\textsuperscript{1}Department of Computer Science, Princeton University\\
\textsuperscript{2}Department of Psychology, Princeton University\\
\textsuperscript{3}Allen School of Computer Science, University of Washington\\
\texttt{\{ryanliu,hyen,raja.marjieh,tomg\}@princeton.edu}\\
\texttt{ranjay@cs.washington.edu}\\
}
\begin{document}

\maketitle

\begin{abstract}

How do we communicate with others to achieve our goals?
We use our prior experience or advice from others, or construct a candidate utterance by predicting how it will be received. However, our experiences are limited and biased, and reasoning about potential outcomes can be difficult and cognitively challenging. 
In this paper, we explore how we can leverage Large Language Model (LLM) simulations to help us communicate better.
We propose the Explore-Generate-Simulate (EGS) framework, which takes as input any scenario where an individual is communicating to an audience with a goal they want to achieve. EGS (1) \textit{explores} the solution space by producing a diverse set of advice relevant to the scenario, (2) \textit{generates} communication candidates conditioned on subsets of the advice, and (3) \textit{simulates} the reactions from various audiences to determine both the best candidate and advice to use. 
We evaluate the framework on eight scenarios spanning the ten fundamental processes of interpersonal communication. 
For each scenario, we collect a dataset of human evaluations across candidates and baselines, and showcase that our framework's chosen candidate is preferred over popular generation mechanisms including Chain-of-Thought.
We also find that audience simulations achieve reasonably high agreement with human raters across $5$ of the $8$ scenarios.  
Finally, we demonstrate the generality of our framework by applying it to real-world scenarios described by users on web forums. 
Through evaluations and demonstrations, we show that EGS enhances the effectiveness and outcomes of goal-oriented communication across a variety of situations, thus opening up new possibilities for the application of large language models in revolutionizing communication and decision-making processes.
\end{abstract}

\section{Introduction}

We communicate with others in order to achieve our goals: to make friends, to accomplish tasks, or simply to convey our intentions~\citep{grice1975logic,sperber1986relevance}. However, it can be hard to find the right words to achieve those goals. Consider a scenario where you are trying to get a discount on an item by haggling with its vendor. There are many strategies that you could use to gain an edge, including complimenting the item, offering to buy multiple items for a discount, or even describing your financial situation and asking them to take pity. With so many potential options, it's difficult to correctly decide which strategy to choose. This problem is not confined to bargaining—everyday communication requires us to make choices about what approaches to adopt, whether we are making friends, impressing others, or navigating romantic conflicts. 

Given a communication scenario, 
how do we decide which strategies to employ? Often, we rely on heuristics such as our prior experience~\citep{schacter2007remembering} or on advice we receive from others~\citep{yaniv2004receiving}. When we have more time to make careful decisions, we may even play out possible candidates in our minds, simulating the reaction of an imaginary listener and using their imagined reaction to guide our choice~\citep{atance2001episodic}. This idea is formalized in the Rational Speech Act (RSA) model~\citep{goodman2016pragmatic}, which explains people's communication choices in terms of speakers simulating listeners as rational interpreters of possible candidate utterances. 
However, both our experiences and the advice of others are biased by the information we are exposed to, making our heuristics and simulations imperfect and resulting in suboptimal communication outcomes~\citep{gilbert2007prospection}. 
Moreover, reasoning about others' potential reactions can be time-consuming and cognitively challenging~\citep{gilbert1988cognitive}.

\begin{figure}[t]
    \centering
    \includegraphics[width=0.9\linewidth]{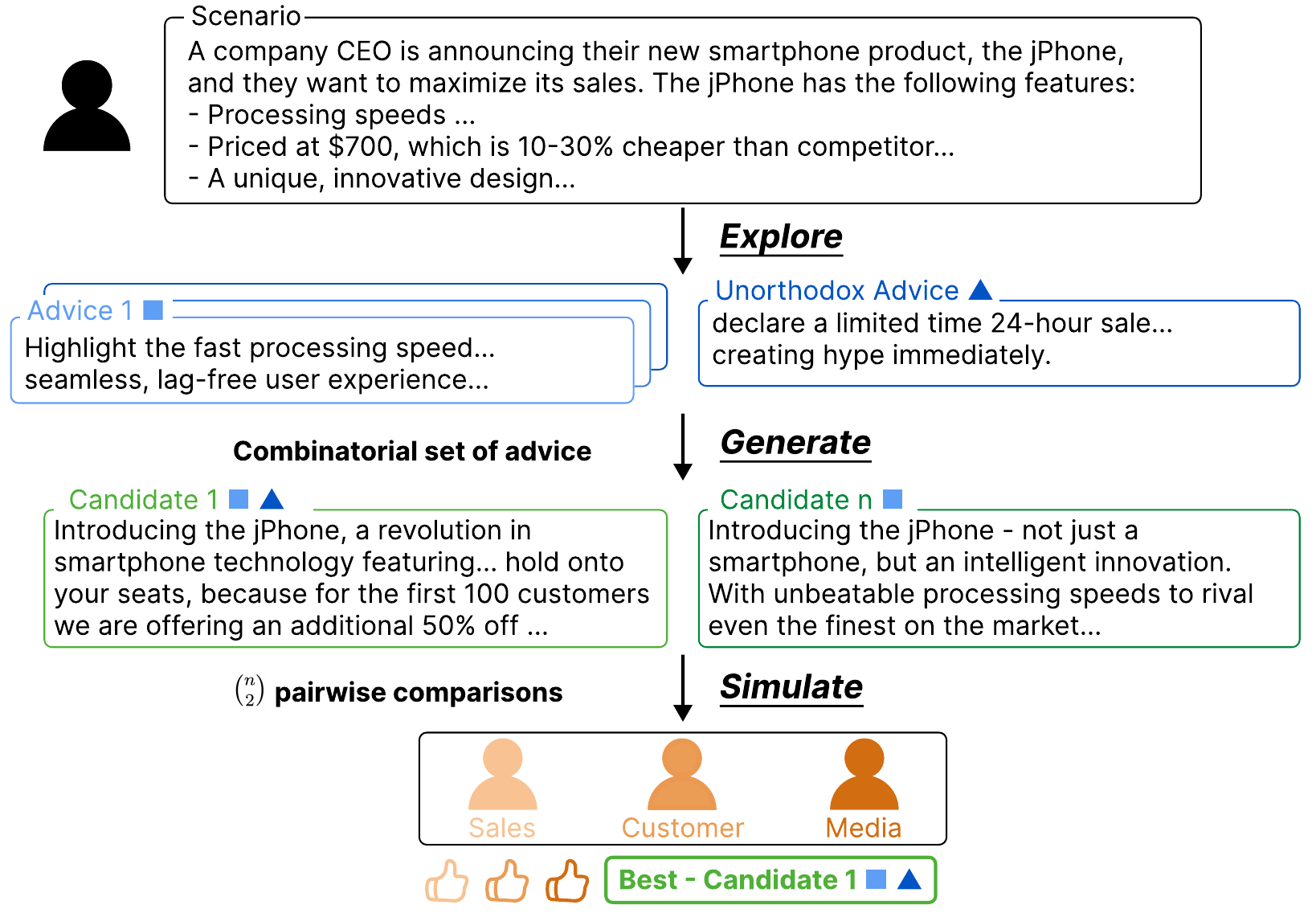}
    \caption{Given a description of a scenario and a goal, our framework generates the best candidate message by simulating advice and stakeholders using an LLM. It \textit{explores} potential pieces of advice that might help, \textit{generates} candidates by conditioning on subsets of the generated advice, and \textit{simulates} audience members who evaluate the various candidates.
    }
    \label{fig:overview}
\end{figure}

Inspired by the newfound capacity of large language models (LLMs) to simulate agents~\citep{Park2023GenerativeAI}, we propose the Explore-Generate-Simulate (EGS) framework, which supports people in exploring communication strategies and developing message candidates while offloading the cognitively challenging simulation of audience reactions. 
More precisely, given an arbitrary communication scenario, EGS first \textit{explores} the space of possible responses by using an LLM to produce both normal and creatively unorthodox advice relevant to the scenario.
Next, it \textit{generates} communication candidates by conditioning an LLM on various subsets of the advice. 
Finally, it \textit{simulates} the reception of each candidate by instructing the LLM to take on the perspectives of possible audiences. Using these simulations, we can determine which candidates and advice are best suited for achieving the communicator's goal. 
By construction, our framework also lets us study whether LLMs, conditioned on the scenario and audience description, can effectively simulate audience reactions.

To evaluate this framework, 
we construct eight diverse scenarios that span the ten fundamental processes of interpersonal communication~\citep{Berger2019Interpersonal} and a variety of communication modalities, relationships, and settings (see Appendix~\ref{app:ten_processes} and \ref{app:scenarios}). The scenarios include an airline representative speaking to the press after a plane crash, a college student trying to write their profile on a dating app, and even everyday situations like a barista interacting with a customer. 
We collect human judgments on the effectiveness of each candidate message and compare how EGS performs against non-simulation baseline methods, including GPT-4 zero-shot and Chain-of-Thought (CoT).
Finally, we analyze the agreement between humans and simulated audiences using real world scenarios drawn from the Stanford Human Preferences (SHP) dataset~\citep{pmlr-v162-ethayarajh22a}.

In both evaluations, we find that our approach convincingly outperforms their respective baselines. We also find relatively high agreement between EGS simulated preferences and human scores across three analyses. 
In ablations, we show that simulated audience evaluations select candidates that are rated higher by humans, and that unorthodox advice improves the quality of generated communication candidates.
We also provide visualizations of qualitative examples, showing LLM-generated messages at each step and highlighting our design choices.
Viewing LLMs as a library of shared experiences and opinions, our simulation approach draws on this library to integrate cultural and individual experience to ultimately help people communicate better.

\section{Related work}
\label{sec:related_work}

Our work uses scenarios motivated by social psychology literature on \textbf{interpersonal relationships}. Our framework utilizes LLM \textbf{agent simulations} as an alternative for \textbf{simulations in the human mind}, and we evaluate it using \textbf{human preferences}.

\textbf{Interpersonal relationships.}
Research in social psychology views interpersonal communication from the perspective of ten fundamental processes that underlie social interaction~\citep{Berger2019Interpersonal}.
These ten processes may be present regardless of the social context within which interpersonal communication occurs 
(see Appendix~\ref{app:ten_processes} for a detailed summary).
Grice identified a set of maxims surrounding the problem of how a cooperative speaker should choose what to say in everyday communication, such as truthfulness or relevance~\citep{grice1957meaning, grice1975logic, grice1989studies}. 
Classical formal models of communication view cooperative communication as information transfer between speaker and listener~\citep{desaussure1916, lewis1969, shannon1948}, with a focus on aligning the true state between agents as the goal of communication~\citep{stalnaker1978}. The Rational Speech Act (RSA) framework~\citep{frank2012, goodman2016pragmatic} draws on both of these traditions, modeling informative speakers as aiming to reduce the listener's uncertainty over the true world state, assuming listeners make rational inferences from the utterances they hear. 

\textbf{Simulations in the human mind.} 
The act of projecting oneself into the future to pre-experience an event is formalized in cognitive science as ``episodic future thinking''~\citep{atance2001episodic}. At the neuropsychological level, brain regions traditionally associated with memory are similarly engaged when people imagine future experiences~\citep{schacter2007remembering}. \citet{szpunar2010episodic} covers a body of work demonstrating a close relation between episodic future thought and the ability to remember personal episodes from one's past. In cognitive psychology, \citet{klein2010facing} find evidence that a goal of long-term memory is to store information about the past to plan for the personal future. Thus, if we consider LLMs as encoding an aggregation of personal experiences across a large subset of human society, they may also have the capacity to simulate epsiodic future thought. Furthermore, as LLMs are theoretically capable to take experiences and infer and represent properties of an agent likely to have had those experiences~\citep{andreas2022language}, they may be able to simulate episodic future thought from the perspective of an average agent with a particular set of properties, leading to the possibility of realistic LLM-simulated audiences. 

\textbf{Agent simulations.} Simulations using large language models are currently seen as an opportunity to expand research in computational social science~\citep{ziems2023can}. 
Although rule-based simulations have traditionally been used to study social phenomena, they are limited by their expressivity~\citep{schelling1971dynamic,easley2010networks}. 
LLMs could potentially simulate more complex social interactions that are harder to codify. They can also be explicitly conditioned to simulate individuals with their own goals and objectives~\citep{jones2022capturing,koralus2023humans, liu2023reviewergpt}.
Recently, this capacity of LLMs was used to simulate social computing systems~\citep{park2022social},
rollout their members’ interactions~\citep{Park2023GenerativeAI}, generate public opinion~\citep{chu2023language},
and even produce individualized subjective experience descriptions~\citep{argyle2023out}.
However, there are uncertainties when using LLM simulations because of their unpredictability~\citep{salganik2006experimental}.
Current LLMs exhibit higher homogeneity of opinions
than humans~\citep{argyle2023out,santurkar2023whose}, and 
combining LLMs with true human samples
is essential to avoid an algorithmic monoculture, leading to collapsed generations that constitute only a
limited perspectives~\citep{kleinberg2021algorithmic,bommasani2021opportunities}.

\textbf{Human preferences.} 
Recent LLMs are trained to following instructions with reinforcement learning using human feedback (RLHF)~\citep{ouyang2022training}. 
To train the reward model in RLHF, they collect human preference data from a team of human annotators~\citep{ouyang2022training, touvron2023llama, bai2022training}.
While previous work use human preference data to train reward models that enable RLHF, we leverage them to test the ability of our model to adapt to scenarios that internet users have encountered~\citep{pmlr-v162-ethayarajh22a}.


\section{The Explore-Generate-Simulate (EGS) Framework}

The EGS framework takes as input an arbitrary scenario where an individual is communicating to an audience with a goal they want to achieve. As output, it suggests a candidate message and set of advice to help the individual achieve their goal. 

\textbf{Example.} An example input may be: 

\begin{quote}
    \textit{A company CEO is announcing their new smartphone product, the jPhone, and they want to maximize its sales. The jPhone has the following features \dots Right now, they are about to give a quick 30-second presentation about the jPhone, broadcasted to major television channels.}
\end{quote}

Here, the individual is the CEO, and their goal is to \textit{maximize the sales of the jPhone}. Given the input, we perform three steps: 

In the \textbf{Explore} step, EGS asks an LLM to generate a diverse set of advice for the scenario. For example, it might suggest: 
\begin{quote}
    \textit{Highlight the fast processing speed and seamless, lag-free user experience.}
\end{quote}

In the \textbf{Generate} step, for each set of advice, EGS asks an LLM to generate candidates for the communication. For the advice above, it might generate: 
\begin{quote}
    \textit{Introducing the jPhone. With unbeatable processing speeds to rival even the finest on the market $\dots$}
\end{quote}

In the \textbf{Simulate} step, EGS first asks an LLM to generate a list of stakeholder audience profiles, each with a unique description and perspective. For the above scenario, a stakeholder might be \textit{media outlets}, and their perspective:
\begin{quote}
    \textit{Your job is to listen to the CEO's presentation, understand the key features and selling points of the smartphone, and relay the information to the public through your media platform $\dots$}
\end{quote}
EGS uses these simulated audience perspectives and evaluates each candidate with each simulated audience member. The metric of the evaluation is the likelihood and magnitude to which the candidate achieves the communicator's goal. We aggregate across audiences to determine the best candidate.

\textbf{Why is default LLM prompting insufficient?} 
When approaching such a problem with LLMs using current methods such as Chain-of-Thought, generated candidates often lack diversity in their wording and approach. Furthermore, the communicator might lack the perspectives of the important audiences in order to accurately evaluate one message candidate against another. Additionally, as we become able to broaden the search space and consider many potential approaches, the communicator can get easily overwhelmed when deciding between all the options, requiring us to build a more scalable method of making judgements between potential message candidates. 

Our EGS framework addresses each of these challenges, and it is modular such that each step can be implemented flexibly based on one's specific scenario. We now summarize our general implementation of each \textit{Explore-Generate-Simulate} component.

\subsection{Explore}
The main purpose of the \textit{Explore} step is to expand the space of possible candiate generations.
This stage generates a list of distinct pieces of advice to later condition the candidate generation upon. 
We following existing literature in Social Psychology, which finds that people recall useful advice~\citep{yaniv2004receiving} or prior experiences~\citep{schacter2007remembering} when considering their next action.
Similarly, our explore stage generates relevant pieces of advice that will be useful for the next stage of the framework.

Additionally, EGS prompts the LLM to generate pieces of advice that are ``unorthodox but potentially helpful for [the agent's] task and goal'' as a method to increase the diversity of candidates.
In the example above, GPT-4 generates the unorthodox advice: 
\begin{quote}
    \textit{In your 30-second presentation, declare a limited time 24-hour sale where the first 100 customers get the phone at an additional 50\% off, creating hype and urgency to buy immediately.}
\end{quote}
Qualitatively, we find that unorthodox advice generated by the model are clever and creative (see Section~\ref{demo:explore}), while quantitatively, it improves the downstream message candidates in 4 out of 8 scenarios (see Section~\ref{sec:ablation_unorthodox}). 
We theorize that this is because LLMs are exposed to a large trove of human experiences through text, allowing it to catch successful strategies that are only employed by smaller communities or even just a few individuals.

\subsection{Generate}
The \textit{Generate} step seeks to create reasonable candidates for the communication guided by advice from the Explore step. Specifically, EGS forms combinatorial subsets of the available advice generated from the previous step. Each subset is used to condition the generation of a candidate message. To account for randomness from the LLM sampling temperature,
we generate three candidates for each advice set.

Following prior work~\citep{Park2023GenerativeAI}, we utilize the ``inner voice'' of the communicator to condition generated message candidates on their assigned advice set, as it makes the LLM more likely to treat the statement as a directive. In particular, we frame the advice set as what the individual remembers in the moment: 
\begin{quote}
    \textit{You remember a few pieces of advice: $\dots$ You decide to focus on using these pieces of advice during your presentation.}
\end{quote}
Conditioning on a combinatorial spread of the advice options further expands the explored solution space. In particular, this allows a message candidate to incorporate many orthogonal advice concepts, potentially leading to better performance. In an ablation we find that though allowing multiple advice does not always improve the average candidate, it does improve the scores of top candidates, being responsible for the highest-scoring candidate in 5 out of the 8 scenarios (see Appendix~\ref{multiple_advice}). 

A key feature that enables the combinatorial advice options is the scalability of EGS. Using LLMs, we are able to generate and compare between candidates at a scale previously unthinkable to human decision-makers, allowing for more sophisticated explorations around communication choices than before. Furthermore, we showcase the quality of these generated candidates through both a demonstration (Section~\ref{demo:generate}) and quantitative analyses (Section~\ref{sec:human_evaluations}).

\subsection{Simulate}
The Simulate step consists of two parts. First, EGS asks the LLM to generate a list of key audiences who have influence over whether the individual accomplishes their goal, and construct profile descriptions for each audience. Then, we ask the LLM to simulate the reactions of these audiences to each candidate message, and aggregate the results to determine which candidate is best for achieving the communicator's goal. 

For each audience profile, we ask the LLM to construct 1) a description of the scenario and their reception of a candidate message from their point of view, and 2) the appropriate question to ask them for how their reception of each candidate would directly influence the communicator's goal. For the generated audience \textit{media outlets} in the example, the question generated for the audience is: 
\begin{quote}
    \textit{In which scenario would you be more likely to give more media coverage and promotion towards the jPhone?}
\end{quote}

We allow the LLM to produce up to $X$ audiences to consider based on the scenario, where $X$ is a user-specified hyperparameter. 
When generating the audiences, we also ask the LLM to provide weights assigned to each audience based on their relative importance in the scenario. 
Using these weights, we aggregate audience evaluations and get the best candidate and advice using a simple weighted sum. For more details on the audience generation, we refer the reader to Appendix~\ref{app:stakeholders}. 
Motivated by literature on cognitive choices \citep{gates2020how}, we provide a list of alternative audience aggregation options and a brief analysis in Appendix~\ref{app:aggregation}. 

Since we are comparing communication candidates that are often very close in quality, LLMs will often lack granularity when giving ratings to individual candidates~\citep{qin2023large}. Instead, we ask LLM-simulated audiences to rate candidates using pairwise comparisons. 
Specifically, we provide the simulated agent with two scenarios, one representing each candidate in the pairwise comparison, and ask it to reason about which is better. 
In some cases, the response will insist that the choice depends on the audience's personal preference, in which we label the outcome as ``tied''. In cases where the output is not in the proper answer format, we ask an LLM to classify the output into one of the pairwise comparison outcomes $o \in$ \{``prefer scenario 1'', ``prefer scenario 2'', ``tie''\}. 

Once we have pairwise comparison outcomes for each audience, we aggregate the scores across audiences to get the best candidate $c^*$:
\begin{align}\label{eq:pairwise}
    c^* = \max_{c} \sum_{c' \neq c} compare(c, c') \qquad
    compare(c, c^\prime) = 
    \begin{cases}
    1 & \text{if ``prefer } c \text{ ''}\\
    0.5 & \text{if ``tie''}\\
    0 & \text{if ``prefer } c^\prime \text{ ''}
    \end{cases}
\end{align}

Intuitively, since each candidate is compared to all other candidates, this method chooses the candidate that performs the best against other candidates. 

\section{Demonstrations}
\label{demo}
\begin{table}[h]
\centering
\caption{A complete look at the ``White Lie During Date'' scenario. The \textbf{Simulate} step uses the date's perspective to compare between the two candidates shown. In the comparison, the left candidate in the \textbf{Generate} step is Scenario 1 and the right candidate is Scenario 2. }
\vspace{10pt}
\label{tab:demo_table}
\resizebox{\textwidth}{!}{%
\begin{tabular}{@{}llll@{}}
\toprule
\multicolumn{4}{l}{\textbf{Scenario: White Lie During Date}} \\ \midrule
\multicolumn{4}{l}{\begin{tabular}[c]{@{}l@{}}
A teenage guy is currently out on a date with a girl at the mall. He is attracted to the girl and wants to make the date \\ 
as enjoyable for her as possible. They are currently in a clothing store and he is waiting while his date is trying out \\ 
an outfit in the changing room. She comes out of the changing room looking very pleased while wearing a black \\
denim jacket, strikes a pose, and asks, "what do you think!". Secretly, he thinks the jacket doesn't look good on her \\
at all, but decides he should try to compliment it anyway. Right now, he is wondering what to say.
\end{tabular}} \\ \midrule
\multicolumn{4}{l}{\textbf{Explore}} \\ \midrule
\begin{tabular}[c]{l@{}}
\textbf{Normal advice 1}\vspace{1pt}\\ 
Share an affirmation that \\ 
re-establishes your interest \\ 
in her, such as ``That style \\ 
can be challenging to pull \\ 
off, but you certainly can \\ 
make anything look good.''
\end{tabular} & \begin{tabular}[c]{@{}l@{}}
\textbf{Normal advice 2}\vspace{1pt} \\ 
Provide a suggestive \\ 
compliment on the jacket, \\ 
for instance, ``Wow! \\ 
That's an edgy look for \\ 
you. It's fun seeing you \\ 
try out new styles.''
\end{tabular} & \begin{tabular}[c]{@{}l@{}}
\textbf{Normal advice 3}\vspace{1pt}\\ 
Compliment elements surrounding \\ 
the jacket, like saying, ``That jacket \\ 
really brings out the color in your \\ 
eyes,'' so you avoid commenting on \\ 
the jacket's overall look yet keep \\ 
the comment positive and about her.
\end{tabular} & \begin{tabular}[c]{@{}l@{}}
\textbf{Unorthodox advice}\hspace{-11pt}\vspace{2pt}\\ 
Compliment the joy \hspace{-11pt}\vspace{0pt}\\ 
the jacket brings \\her, rather than the \\ 
jacket itself.
\end{tabular} \\ \midrule
\multicolumn{4}{l}{\textbf{Generate}} \\ \midrule
\multicolumn{2}{l}{\begin{tabular}[c]{@{}l@{}}\textbf{Normal advice 3 + Unorthodox advice} \vspace{3pt} \\ 
Wow, I love how that jacket brings out the sparkle in \\
your eyes, and I can see how happy it makes you feel. \\
It certainly adds a unique flair to your whole look.
\end{tabular}} & \multicolumn{2}{l}{\begin{tabular}[c]{@{}l@{}}\textbf{Normal advice 1 + Normal advice 2}\vspace{2pt} \\ 
Wow! That's an edgy look for you, it really shows off your \hspace{-11pt}\vspace{0pt}\\ 
daring side. You know, that style can be a bit tricky to pull \hspace{-11pt}\vspace{0pt}\\ 
off, but it seems like there's nothing you can't make look \\ 
good. I'm having a blast seeing you try out new styles.
\end{tabular}} \\ \midrule
\multicolumn{4}{l}{\textbf{Simulate}} \\ \midrule
\multicolumn{4}{l}{\begin{tabular}[c]{@{}l@{}}
The compliment and feedback I prefer will largely depend on my personality and what I value from my date's response. \hspace{-12pt}\vspace{3pt}\\ 
In Scenario 1, he gave a straightforward compliment, focusing on the effect the jacket has on my physical appearance, \hspace{-11pt}\vspace{0pt}\\
particularly my eyes, and how it elevifies my happiness. He acknowledges my feeling of joy and captures the \\ 
personalized aspect of my style, which I really appreciate. It shows he's paying close attention to 
me not only at \\visual but also at emotional level. \vspace{4pt} \\ 
In Scenario 2, he described how the jacket contributes to my overall style, suggesting I'm daring and edgy. He admires \hspace{-14pt}\vspace{0pt}\\ 
how I pull off tricky styles, which indicates he thinks I have a good sense of fashion. However, the phrase ``it seems \hspace{-11pt}\vspace{0pt}\\
like there's nothing you can't make look good'' might come off as a generic compliment. But the excitement he's \\
showing in wanting to continue watching me try new styles implies he's also interested in sharing the shopping \\
experience together. \vspace{4pt} \\ 
Considering all points, while both scenarios provide complimentary feedback, I prefer Scenario 1. It resonates well as \hspace{-11pt}\vspace{0pt}\\ 
he not only notices the physical aspects but also the happiness it brings me, which very well shows he cares about my \hspace{-11pt}\vspace{0pt}\\ 
feelings. \vspace{4pt} \\ 
\textbf{Answer:} Scenario 1
\end{tabular}} \\ \bottomrule
\end{tabular}%
}
\end{table}

In Table~\ref{tab:demo_table}, we showcase a complete run of the EGS framework on one of our scenarios, using GPT-4 as the LLM. The scenario describes a dilemma that a teenager faces during a date at the mall. 

\paragraph{Explore}
\label{demo:explore}

In the \textit{Explore} step, we generate three normal advice\footnote{In particular, we note that ``suggestive compliment'' in advice 2 should be interpreted as a compliment that hints at a subtler message, i.e., that the jacket doesn't look great, and not as a compliment with sexual connotations. } and one unorthodox advice based on the scenario. 
In the demo, we observe that each advice generated takes into account the young man's true impression, and attempts to bridge the gap between his impression and his date's expectation. More generally, we find that GPT-4 tends to avoid using or suggesting the use of lies, which we believe is an importantly aligned value when trying to improve interpersonal communication. 

The unorthodox advice suggested provides a unique strategy for bridging between the young man's impression and his date's expectation by suggesting for him to focus on the joy that the jacket makes her feel. Though this may be more niche and unexpected than the normal advice options, it fits well with the scenario and is also natural to implement. Across the scenarios, we found that the unorthodox advice suggested were both reasonable and creative, surpassing our expectations and demonstrating the value that EGS can bring into everyday communication problems. 

\paragraph{Generate}
\label{demo:generate}
In the \textit{Generate} step of the demo, we highlight two candidates, each generated using two pieces of advice. In the first candidate, we observe the adoption of the normal advice with ``that jacket really brings out the sparkle in your eyes'', followed by the use of the unorthodox advice `` I can see how happy it makes you feel''. We find that in general, GPT-4 often adapts or embellishes the advice, such as adjusting ``color'' to ``sparkle'' to presumably account for the black color of the jacket, while also matching ``sparkle'' with ``how happy it makes you feel''. 

In the second candidate, we observe an interleaving of the advice, where the candidate starts off with ``That's an edgy look for you'' from advice 2, moves to ``that style can be a bit tricky to pull off'' from advice 1, and returns to advice 2 with ``I'm having a blast seeing you try out new styles''. More generally, we observe that the \textit{Generate} step is often flexible on the structure on how it incorporates the advice, while seamlessly connecting them with each other and the scenario context. 
We also note that even though our framework prompt structure stays the same, the candidates always align with the scenario, speaker, and goal, which we consistently observe across all our experiments.

\paragraph{Simulate}
In the \textit{Simulate} step of the demo, we provide one of the comparisons from the simulated date's perspective between the two candidates shown in the \textit{Generate} step. 
The simulated audience starts off by individually reasoning about each scenario, before composing them into a final verdict. 
This verdict is eventually combined with all others to select the best response/advice. 

For Scenario 1, the simulated reaction starts off with a summary of the candidate: ``a straightforward compliment, focusing on the effect the jacket has on [\dots] my eyes, and how it elevifies my happiness''. Then, it moves to what this means for the date: ``He acknowledges my feeling of joy and captures the personalized aspect of my style''. Lastly, it synthesizes this into a higher-level reflection about the situation itself: ``It shows he's paying close attention to me not only at visual but also at emotional level''. In general, we find hierarchical reasoning to be common in these analyses, allowing the simulated comparisons to consider many facets of the communication candidates. 

For Scenario 2, though the simulated reaction proceeds sentence-by-sentence, there exists a two-level hierarchy in the reaction to the last sentence, highlighting her date's excitement before concluding that he is ``interested in sharing the shopping experience together''. The simulated reaction also calls out the phrase ``there's nothing you can't make look good'' as generic, which is an interpretation that is very reasonable but likely not universal. In cases like this where the audience is underspecified and their reaction is unpredictable, we find that the LLM navigates this space quite smoothly, often adopting a reaction that represents the majority of the audience demographic. We believe that this is a large part of why \textit{Simulate} is able to achieve a high agreement with human raters as well as online users, which we show with experiments in the next two sections.

\section{Human evaluations}
\label{sec:human_evaluations}
We collect human evaluations for all candidates from the \textit{Generate} step and two baselines in each scenario, and use this to evaluate 1) How our framework's chosen advice and candidate performs relative to the others, and 2) The amount of agreement between human raters and GPT-4 pairwise comparisons. 

\textbf{Data collection.}
We construct eight diverse scenarios that span the ten fundamental processes of interpersonal communication, listed in Appendix~\ref{app:ten_processes} and \ref{app:scenarios}. 
For each scenario, we collect human ratings on all candidates from the \textit{Generate} step of EGS, two baselines (GPT-4 zero-shot and CoT), and two sets of ablated candidates generated by altering the search space of the \textit{Explore} step of EGS. 

Within EGS, we use three pieces of normal advice and one piece of unorthodox advice, and combine them combinatorially. Candidates are conditioned on advice sets containing up to two pieces of advice, which we decided via a preliminary study (see Section~\ref{sec:ablation_num_advice}), yielding 10 distinct advice sets. We generate three candidates for each set to form a total of 30 candidates. 

In the GPT-4 zero-shot baseline, we use the advice generation prompt, but replace the request to generate advice with directly asking what it would say if it were the agent in the scenario. In the GPT-4 Chain-of-Thought baseline~\citep{wei2022chain}, we use the same format, but ask the model to generate its reasoning about the scenario before providing what it would say. We also generate three candidates for each baseline.

For the two altered \textit{Explore} versions of EGS, we prompt the model to generate advice that are encouraging or irrelevant respectively, rather than conceptual. We use three pieces of advice each, resulting in 6 advice sets or 18 candidates per version. 

Human ratings were on a 0-10 Likert scale, with (0) representing a highly negative impact, (5) representing a relatively neutral impact, and (10) representing a highly positive impact on the communicator's goal. To illustrate, for the scenario shown in the demonstration (Section~\ref{demo}):\vspace{2pt} \\
(0) \hspace{5pt}``I think his comment would make me enjoy the date a lot less.'' \\
(5) \hspace{5pt}``I think his comment would not affect how much I am enjoying the date.'' \\
(10) ``I think his comment would make me enjoy the date a lot more.''

\textbf{Participants.}
Our behavioral dataset comprises $12,180$ human judgments from $N=652$ UK participants crowdsourced using Prolific\footnote{\url{https://www.prolific.co}}. 
All participants provided informed consent prior to participation in accordance with an approved institutional review board (IRB) protocol, and were paid at a rate of 12 USD per hour. Overall, we collected $20$ judgments per candidate and baseline for each of the eight experimental scenarios, and each participant provided up to $20$ judgments. This yielded an excellent average inter-rater reliability (IRR) of $r=.82$ computed using a split-half correlation method with Spearman-Brown correction~\citep{brown1910some}. For more details on the data collection procedure, we refer the reader to Appendix~\ref{app:experiment_procedure}.

\subsection{Result: EGS outperforms GPT-4 Zero-shot and CoT}
\label{sec:result_main}
\begin{table}[]
\centering
\caption{The best candidate message selected by EGS outperforms GPT-4 zero-shot in human ratings across all constructed scenarios, and outperforms GPT-4 with CoT in five scenarios and a subset of a sixth scenario. 
}
\vspace{10pt}
\label{tab:baseline_comparison}
\resizebox{0.82\textwidth}{!}{%
\begin{tabular}{@{}lccc@{}}
\toprule
Scenario                        & GPT-4 zero-shot & Chain-of-Thought & EGS (ours)    \\ \midrule
Plane Crash                     & 6.83           & 5.98             & \textbf{7.95} \\
Product Launch                  & 5.73           & 5.95             & \textbf{7.05} \\
Bargaining                      & 4.68           & \textbf{5.98}    & 5.85          \\
Bargaining (-unorthodox advice) & 4.68           & 5.98             & \textbf{6.60} \\
Barista                         & 5.58           & \textbf{5.78}    & 5.40          \\ 
Sharing Secrets                 & 3.67           & 4.17             & \textbf{5.55} \\
Dating App                      & 5.42           & \textbf{6.48}    & 5.05          \\
White Lie During Date           & 6.12           & 6.02             & \textbf{6.70} \\
Marriage Argument               & 6.78           & 6.70             & \textbf{7.80} \\ \midrule
Average                         & 5.50           & 5.89             & \textbf{6.44} \\ \bottomrule
\end{tabular}%
}
\end{table}
We compared the average human ratings of the candidate selected by EGS and two baselines, GPT-4 zero-shot and GPT-4 Chain-of-Thought. EGS outperforms GPT-4 zero-shot in all scenarios, and Chain-of-Thought in five scenarios (Table~\ref{tab:baseline_comparison}). 
On average, EGS outperforms GPT-4 zero-shot by $0.94$ ($17.1\%$) and Chain-of-Thought by $0.55$ ($9.3\%$). In four of the eight scenarios, EGS outperforms both baselines by at least $1$ full point.

We also note that all outputs from EGS surpassed a mean score of 5, indicating that they all had a positive impact on the communicator's goal, whereas this was not the case for either baseline. This suggests that EGS was able to (1) expand the search space such that candidates with positive outcomes are generated, and (2) correctly identify which of the generated candidates would actually lead to positive outcomes using audience simulations. 

Furthermore, in the Bargaining scenario where Chain-of-Thought outperforms EGS (5.98 vs. 5.85), we find a large discrepancy between human and GPT-4 preferences on the unorthodox piece of advice (see Section \ref{sec:agreement_winlose_scenario} for details). After reducing the Explore space by removing the candidates generated from this piece of advice, EGS outperforms Chain-of-Thought by a large margin (6.60 vs. 5.98). 
More generally, we find that the performance of EGS highly correlates with the agreement between GPT-4 and human ratings in the \textit{Simulate} step, which we delve into next.

\subsection{Result: High agreement between human scores and LLM comparisons 
}
\label{sec:agreement}

\textbf{Multilevel model across scenarios.}
Using a multilevel model, we analyze the agreement between GPT-4 and human raters on the communication candidates across the eight different scenarios. Specifically, we determine if candidates preferred by the combined stakeholders had a higher mean score from human raters than those less preferred. 

In the multilevel model, the independent variable was the weighted sum of stakeholder pairwise judgements, where each stakeholder evaluates each pair as better ($1$), worse ($0$), or tie ($0.5$) three times and takes the mean. Our dependent variable was the scores provided by human raters, which were on a $0$--$10$ scale. 
The data has a three-level structure, with individual items nested within pairwise comparisons, which were further nested within the scenarios. The scenarios were treated as random effects to control for inherent differences among scenarios. 

The modeling results revealed a significant fixed effect for the weighted pairwise judgements on the score provided by human raters $(\text{coef} = 0.427, p = 0.041)$, providing strong evidence of significant agreement between the GPT-4 and human evaluations across scenarios. We found that this effect differed across scenarios, indicating the multilevel model's appropriateness in taking into account the hierarchical nature of our data.

\textbf{GPT-4 pairwise winners vs. losers' mean human ratings per scenario.}
\label{sec:agreement_winlose_scenario}
Within each scenario, we conduct a paired samples t-test across the preferred and less preferred candidates of each pairwise comparison, and find that preferred candidates are significantly better in five of the eight scenarios with $\alpha = 0.001$ (see Table~\ref{tab:agreement_tom}).  

We also find that the Bargaining scenario contained a piece of controversial advice that was in both EGS's best advice set and best candidate, but was given a mean human rating of 5.56 (candidates without this advice averaged a score of 6.20, t-test with/without this advice $p < 0.001$): 
\begin{quote}
    ``Consider offering to promote the artisan's work on your social media in exchange for a discount on the vase.'' 
\end{quote}
When we remove this advice and re-run the framework, the difference returns to being highly significant in favor of the preferred candidate in the pairwise comparisons. For our experiments, this version of shown under ``Bargaining (-unorthodox advice)''. 

More generally, for candidates containing unorthodox advice, not only do we find higher disagreement between human raters and EGS, we also find higher disagreement within human raters themselves in seven of the eight scenarios (see Appendix~\ref{app:irr}). This reflects that simulating audience reactions in the less common situations created by unorthodox advice is less consistent for humans as well as LLMs.

Moreover, though this metric allows us to test for statistical significance, it is largely affected by easier comparisons which have a large disparity in scores, i.e., comparisons where one candidate is clearly better than the other. Thus, we follow with a percentage agreement analysis where each pair is weighted equally.

\textbf{Percentage agreement within individual scenarios.}
\label{sec:percentage_agreement}
\begin{table}[]
\centering
\caption{In both metrics, we find high agreement across GPT-4 and human crowdworkers in five scenarios and a subset of a sixth scenario. 
Preferred and less preferred values are mean scores across all GPT pairwise evaluations, with standard errors of the mean. *, **, and *** denote $p < 0.05$, $p < 0.01$, and $p < 0.001$ respectively. Agreement is the expected percent agreement when randomly breaking ties.}
\vspace{10pt}
\label{tab:agreement_tom}
\resizebox{0.8\textwidth}{!}{%
\begin{tabular}{@{}llll@{}}
\toprule
Scenario & Preferred & Less Preferred & \multicolumn{1}{l}{Agreement} \\ \midrule
Plane Crash & \textbf{6.19 $\pm$ 0.03***} & 5.86 $\pm$ 0.03 & 0.63 \\
Product Launch & \textbf{6.20 $\pm$ 0.03***} & 5.87 $\pm$ 0.03 & 0.67 \\
Bargaining & 5.90 $\pm$ 0.03 & \textbf{5.99 $\pm$ 0.02*} & 0.53 \\
Bargaining (-unorthodox advice) & \textbf{6.35 $\pm$ 0.04***} & 6.06 $\pm$ 0.03 & 0.69 \\
Barista & \textbf{4.66 $\pm$ 0.08***} & 3.53 $\pm$ 0.09 & 0.64 \\ 
Sharing Secrets & \textbf{5.72 $\pm$ 0.03***} & 4.99 $\pm$ 0.04 & 0.78 \\
Dating App & 5.24 $\pm$ 0.03 & \textbf{5.44 $\pm$ 0.03***} & 0.41 \\
White Lie During Date & 6.70 $\pm$ 0.03 & \textbf{6.81 $\pm$ 0.03**} & 0.43 \\
Marriage Argument & \textbf{6.34 $\pm$ 0.04***} & 6.01 $\pm$ 0.03 & 0.65 \\ \bottomrule
\end{tabular}%
}
\end{table}
Agreement is calculated between $n$ mean human ratings and $n\choose{2}$ pairwise comparisons as follows. For each pair of candidates $i$, $j$, we calculate their mean human scores $h_i$, $h_j$, and break ties favoring the candidate with the higher mean score when normalizing within each participant. 
For each simulated audience, we average over $3$ generated comparisons between $i$ and $j$, and then compute a weighted sum across audiences using LLM-assigned weights. From this, we derive three classes - \{majority prefers $i$, majority prefers $j$, balanced\} - denoted $m_i$, $m_j$, and $m_b$ respectively. 
\begin{align}
    \text{agreement} = \frac{1}{{n\choose2}}\sum_{i \neq j} A(i, j) \qquad 
    A(i, j) = 
    \begin{cases}
    1 & \text{if } h_i > h_j, m_i \;\;\textrm{or}\;\; h_i < h_j, m_j\\
    0.5 & \text{if } m_b\\
    0 & \text{otherwise}
    \end{cases}
\end{align}

This metric can be interpreted as the expected percentage agreement between $\{h_i > h_j, h_i < h_j\}$ and $\{m_i, m_j\}$, with $m_b$ tiebreaked randomly.
We do not adjust based on the amount of agreement expected by chance (such as in Cohen's $\kappa$) as the classes of our pairwise ratings are by definition unbiased since they only depend on the order in which the candidates are shown to the rater. In particular, we did not find evidence of any positional bias~\citep{wang2023large}, i.e., preferring Scenario 1 over Scenario 2 regardless of content, in the pairwise comparisons done by EGS (see Appendix~\ref{app:positional_bias}). 

Using this metric, we also find the agreement of between human raters and GPT-4 pairwise evaluations to be reasonably high (Table~\ref{tab:agreement_tom}), with five of the scenarios (plane crash, product launch, barista, sharing secrets, and marriage argument) reaching an agreement above $0.6$. Furthermore, when we only consider more obvious pairwise comparisons, i.e., where the difference in mean human scores between candidates exceeds certain thresholds, we notice that the agreement becomes more pronounced, with four scenarios eventually reaching $0.8$ agreement with moderate sample sizes (See Appendix~\ref{app:large_differences} for details). These match the results from the previous analysis, confirming our findings that GPT-4 is able to simulate human audience judgements to a reasonable degree.

We also find that removing the unorthodox advice in the Bargaining scenario yields a high agreement of 0.69, confirming that EGS aligns with human ratings in the scenario without the unorthodox advice. 
However, for the Dating App and White Lie During Date scenarios, we observe an agreement of less the $0.5$, suggesting that EGS with GPT-4 struggles to accurately evaluate dating advice.

\subsection{Ablations and Additional Analyses}
\label{sec:ablations}

\textbf{Simulated audience evaluations select high-performing candidates. }
We perform an ablation on the effectiveness of the \textit{Simulate} step by comparing the best candidate and advice of EGS against a baseline that selects a random candidate from the \textit{Generate} step. In five scenarios and the bargaining scenario without unorthodox advice, EGS selects a more optimal candidate and set of advice than the baseline (see Table~\ref{tab:ablation_merged}). 

We note that the three scenarios that the \textit{Simulate} step performs poorly at are also the scenarios which had low human-GPT agreement, which matches the intuition that the accuracy of the comparison process heavily determines the performance of the chosen outcome. 
\begin{table}[]
\centering
\caption{An ablation on the \textit{Simulate} step shows that simulating audiences and choosing the best candidate performs better than selecting a candidate or set of advice at random for five scenarios and a subset of a sixth scenario. Standard deviations are shown for both candidates and advice sets. 
}
\vspace{10pt}
\label{tab:ablation_merged}
\resizebox{0.81\textwidth}{!}{%
\begin{tabular}{@{}lcc|cc@{}}
\toprule
Scenario                        & \begin{tabular}[c]{@{}c@{}}Average\\ candidate\end{tabular} & \begin{tabular}[c]{@{}c@{}}EGS\\ candidate\end{tabular} & \begin{tabular}[c]{@{}c@{}}Average\\ advice set\end{tabular} & \begin{tabular}[c]{@{}c@{}}EGS\\ advice set\end{tabular} \\ \midrule
Plane Crash                     & 6.03 $\pm$ 0.66                                             & \multicolumn{1}{c|}{\textbf{7.95}}                      & 6.03 $\pm$ 0.40                                              & \textbf{6.37}                                            \\
Product Launch                  & 6.03 $\pm$ 0.57                                             & \multicolumn{1}{c|}{\textbf{7.05}}                      & 6.03 $\pm$ 0.44                                              & \textbf{6.68}                                            \\
Bargaining                      & \textbf{5.95 $\pm$ 0.54}                                    & \multicolumn{1}{c|}{5.85}                               & \textbf{5.95 $\pm$ 0.43}                                     & 5.63                                                     \\
Bargaining (-unorthodox advice) & 6.20 $\pm$ 0.45                                             & \multicolumn{1}{c|}{\textbf{6.60}}                      & 6.20 $\pm$ 0.31                                              & \textbf{6.57}                                            \\
Barista                         & 4.08 $\pm$ 1.84                                             & \multicolumn{1}{c|}{\textbf{5.40}}                      & 4.08 $\pm$ 1.86                                              & \textbf{5.55}                                            \\ 
Sharing Secrets                 & 5.36 $\pm$ 0.84                                             & \multicolumn{1}{c|}{\textbf{5.55}}                      & 5.36 $\pm$ 0.78                                              & \textbf{5.95}                                            \\
Dating App                      & \textbf{5.41 $\pm$ 0.68}                                    & \multicolumn{1}{c|}{5.05}                               & \textbf{5.41 $\pm$ 0.57}                                     & 4.98                                                     \\
White Lie During Date           & \textbf{6.75 $\pm$ 0.61}                                    & \multicolumn{1}{c|}{6.70}                               & \textbf{6.75 $\pm$ 0.39}                                     & 6.37                                                     \\
Marriage Argument               & 6.18 $\pm$ 0.79                                             & \multicolumn{1}{c|}{\textbf{7.80}}                      & 6.18 $\pm$ 0.62                                              & \textbf{6.88}                                            \\ \bottomrule
\end{tabular}%
}
\end{table}

\textbf{Which search space should we \textit{Explore}, and how much does this choice matter?}
We compare our search over conceptual advice with encouragement-type advice and irrelevant advice across all scenarios, and find that conceptual advice generates better candidates on average for six of the eight scenarios (see Table~\ref{tab:ablation_search_space}). In the encouragement and irrelevant versions, we modify the \textit{Explore} step such that GPT-4 is asked to generate advice that encourages the communicator instead of providing strategic value and advice irrelevant to the scenario, respectively. 

The mean score of human ratings for candidates in the conceptual search space outperforms the irrelevant search space is all eight scenarios, with seven scenarios being highly statistically significant ($\alpha < 0.01$). 
In the plane crash scenario, the irrelevant candidates had an average score of more than 4 points lower than the conceptual candidates on a 10-point scale, demonstrating the importance of the choice of search space. Conceptual advice also outperforms the encouragement search space in six of the eight scenarios, with five scenarios being highly significant ($\alpha < 0.01$). However, conceptual advice also performs significantly worse than encouragement advice in the barista scenario, suggesting that the optimal advice prior might vary based on the specific setting. 
\begin{table}[]
\centering
\caption{An ablation on the \textit{Explore} step shows that searching the conceptual space for advice to condition our candidates upon creates higher-quality candidates than searching for unrelated or encouragement-type advice in six scenarios. Errors shown are standard errors of the mean. *, **, and *** denote $p < 0.05$, $p < 0.01$, and $p < 0.001$ respectively. }
\vspace{10pt}
\label{tab:ablation_search_space}
\resizebox{0.72\textwidth}{!}{%
\begin{tabular}{@{}llll@{}}
\toprule
Scenario              & \begin{tabular}[c]{@{}l@{}}Conceptual \\ Search Space\end{tabular} & \begin{tabular}[c]{@{}l@{}}Irrelevant\\ Search Space\end{tabular} & \begin{tabular}[c]{@{}l@{}}Encouragement\\ Search Space\end{tabular} \\ \midrule
Plane Crash           & \textbf{6.03 $\pm$ 0.07}                                           & 1.92 $\pm$ 0.11***                                                & 3.91 $\pm$ 0.14***                                                   \\
Product Launch        & \textbf{6.03 $\pm$ 0.09}                                           & 5.82 $\pm$ 0.10                                                   & 4.26 $\pm$ 0.12***                                                   \\
Bargaining            & \textbf{5.95 $\pm$ 0.09}                                           & 3.98 $\pm$ 0.14***                                                & 5.00 $\pm$ 0.12***                                                   \\
Barista               & 4.08 $\pm$ 0.11                                                    & 2.85 $\pm$ 0.14***                                                & \textbf{5.84 $\pm$ 0.12***}                                          \\
Sharing Secrets       & \textbf{5.36 $\pm$ 0.09}                                           & 4.91 $\pm$ 0.11**                                                 & 4.89 $\pm$ 0.11**                                                    \\
Dating App            & \textbf{5.41 $\pm$ 0.08}                                           & 4.65 $\pm$ 0.11***                                                & 5.34 $\pm$ 0.11                                                      \\
White Lie During Date & \textbf{6.75 $\pm$ 0.08}                                           & 3.62 $\pm$ 0.13***                                                & 5.57 $\pm$ 0.13***                                                   \\
Marriage Argument     & 5.41 $\pm$ 0.10                                                    & 4.39 $\pm$ 0.14***                                                & \textbf{6.45 $\pm$ 0.13}                                             \\ \bottomrule
\end{tabular}%
}
\end{table}

\textbf{How effective is unorthodox advice?}
\label{sec:ablation_unorthodox}
We compare the human scores of candidates that utilize unorthodox advice with those that do not, and find that those that contain unorthodox advice perform better in 4 of the 8 scenarios, with 2 scenarios having statistically significant differences ($\alpha = 0.05$). We also find that 2 of the scenarios' highest human-rated candidates are conditioned on the unorthodox advice, showing that it has the potential to create better communication options as well. In particular, we note that since the maximum amount of advice per candidate is limited to 2, the unorthodox advice is actively replacing a normal piece of advice when it does better. 

\textbf{Performance decreases with too many pieces of advice.} \label{sec:ablation_num_advice}
One potential benefit for large language models is that it is able to consider a much larger set of advice than humans can. With this in mind, we conducted an initial study with one scenario where candidates could be based on up to three pieces of advice. However, we found a large drop in performance when three pieces of unique advice were incorporated. In particular, we observed a common behavior where GPT-4 attempts to incorporate all of the advice into the generated message body, resulting in utterances that were conceptually disconnected and excessively long. 

Based on this result, we limited the maximum number of advice for any candidate in the \textit{Generate} step to two in our main experiments. This is also intuitively aligned with the applications of the model. In realistic improvisation scenarios, a human agent might find it much easier to effectively use less pieces of advice~\citep{miller1956}. 
However, as the capabilities of LLMs continue to improve, we may see changes in the behavior of these models that may allow for the utilization of many more pieces of advice at once. 
\section{Broader internet user simulation} 
We further evaluate simulated audiences, 
the central method of EGS, on a much broader space of interaction using the Stanford Human Preferences (SHP;~\citealt{pmlr-v162-ethayarajh22a}) dataset\footnote{\url{https://huggingface.co/datasets/stanfordnlp/SHP}}. 
The dataset contains a diverse set of 385K collective human preferences over responses to human-written questions/instructions across 18 subject areas, spanning a wide variety of scenarios in an online forum setting. 
The domains span from cooking to legal advice, making it a robust test bed for the ability of GPT-4 to simulate different types of audiences. 
Each entry contains a forum post/scenario, two comments from the post discussion, and the number of upvotes each comment received from forum users. We provide two examples of the questions/scenarios in Tables~\ref{tab:shp-culinary-example} and~\ref{tab:shp-socialscience-example} in the Appendix. 

We evaluate three different methods on their accuracy of predicting the comment with more upvotes. 
First, as a baseline, we use Chain-of-Thought, which prompts the LLM to reason about the post before deciding on the comment that is likely to have more upvotes.
Next, we use two versions of our EGS Simulate step, each containing one audience profile that we define beforehand. 
The first method, which we call EGS Redditor simulation (Default), conditions the LLM to take the perspective of a Redditor browsing a particular forum before prompting it to reason about which comment it is more likely to upvote; 
The second method, EGS Redditor simulation (Funny), follows the Default setting but specifies that the simulated Redditor is more likely to upvote funny and entertaining comments. Prompts used and examples of simulated audience outputs can be found in Appendix~\ref{app:shp}. 

Following the authors of SHP, we first filter the dataset by a ratio threshold of 3. This ensures that the more preferred comment in each pair is nontrivially preferred over the other comment. 
To reduce the cost of API access, we select $5$ subreddits and use a random sample of $100$ test examples from each, resulting in a total of $500$ evaluations (see Appendix~\ref{app:shp} for more details).

\begin{table}[t]
\centering
\caption{EGS conditioned on different audience prompts outperforms GPT-4 with Chain-of-Thought on redditor preferences. Legaladvice and askculinary are more casual forums, and thus conditioning the simulated audience to be entertainment-seeking leads to the highest performance. In more serious forums such as asksocialscience, the default audience is more accurate. }
\vspace{10pt}
\label{tab:shp-results}
\resizebox{0.9\textwidth}{!}{%
\begin{tabular}{@{}lccc@{}}
\toprule
Domain           & \multicolumn{1}{l}{\begin{tabular}[c]{@{}l@{}}Chain-of-Thought\end{tabular}} & \begin{tabular}[c]{@{}c@{}}EGS Redditor Simulation\\ (Default)\end{tabular} & \begin{tabular}[c]{@{}c@{}}EGS Redditor Simulation\\ (Funny)\end{tabular} \\ \midrule
legaladvice      & 71.0                                                                            & 70.0                                                                        & \textbf{76.0}                                                             \\
askculinary      & 59.0                                                                            & 60.0                                                                        & \textbf{70.5}                                                             \\
askhr            & 72.0                                                                            & \textbf{76.0}                                                               & 72.5                                                                      \\
eli5             & 74.0                                                                            & \textbf{76.0}                                                               & 71.5                                                                      \\
asksocialscience & 77.8                                                                            & 79.4                                                                        & \textbf{61.9}                                                             \\ \bottomrule
\end{tabular}%
}
\end{table}

We present a comparison between CoT and the two simulated audiences in Table~\ref{tab:shp-results}. 
We observe a common trend that EGS Redditor simulation is equal to or better than the CoT baseline, suggesting that the model is able to make decisions more aligned with the real users (i.e. the Redditors who browsed and upvoted the comments) when explicitly prompted to simulate them.
Furthermore, directing the model to look for funny and entertaining comments can significantly boost the performance on more casual forums such as cooking and legal advice. 

In domains such as asksocialscience\footnote{\url{https://www.reddit.com/r/AskSocialScience/}, the forum's rules (at the time of submission) include ``all claims in top level comments must be supported by citations to relevant social science sources" and ``comments must be serious attempts to answer the question".} where strict rules are enforced on the informativeness and sincerity of comments, we expect the demographic of the forum audience to match Redditor Simulation (Default) more, and the performance of EGS Redditor simulation (Funny) drops accordingly. 
Our experiment further validates that the \textit{Simulate} component can generalize to diverse internet user data, and demonstrates that its performance can be further boosted with a better understanding of the corresponding audience.

\section{Discussion}

In the following discussion, we highlight benefits, extensions, domain-specific applications, potential limitations, and broader impacts of our work. 

\subsection{Interpretable explanations and immediately available alternatives}
One benefit of EGS is that it is able to provide easily accessible and interpretable explanations for its judgements. As each comparison between candidates is justified with detailed reasoning, a user can easily find explanations for why one candidate is preferred over another from the perspective of any simulated audience. Combined, these can form a collective explanation for how the framework chose the best message candidate, or the ranking of any other candidate in particular. 

Similarly, if a user dislikes EGS's recommended candidate or advice, they can easily access a list of alternatives that also performed well in the pairwise comparisons, or consider changing the weights of the stakeholders or the aggregation mechanism (alternatives provided in Appendix~\ref{app:aggregation}) based on their preferences. 

\subsection{Extending the framework with user controls}

While we focus on having the LLM generate potential audiences, as mentioned above, there may be situations where an end-user might want to control the audience members and their weights more explicitly. 
For example, if an airline company did want to put together a statement about a crash, and knew that their primary audience was ``family members'' and a ``regulatory board'' that determines what statements are not permitted, they can insert one or more custom designed audience members into the framework, or edit an existing audience profile to their liking. Similar kinds of user controls can limit or change the pieces of advice used within EGS, the types of candidates generated, and even the outcomes of the simulated comparisons.

Taken further, this also enables a human-in-the-loop version of EGS where any of the steps can be fully reviewed and/or amended by a human viewer. We expect this option to be structured as a trade-off between performance and ease of use, where a paradigm that incorporates more human intervention into EGS may see an increase in quality and alignment at the cost of increased human effort. 

\subsection{Extending to multi-turn conversations}

While EGS primarily involves the optimization of a single message, it can be extended to simulate multi-turn dialogues of arbitrary length by simply extending the \textit{Generate} step to include one or more responses from the audience and repeating for the duration of the dialogue (visually, the acronym becomes EGG$\dots$GS). At the end, we can \textit{Simulate} the audiences' reactions to the entire chain of generated dialogue and compare them to find the best candidate at each utterance opportunity and the best overall advice. 

One can also envision a variant where the search space is re-explored at every single \textit{Generate} step (visually, EGEG$\dots$EGS). This allows the advice to change based on the flow of the conversation, which could potentially lead to better performance but may not be realistic for live communication scenarios. 

A limitation in the above approaches is that they generate an exponential amount of potential candidates with respect to the length of the dialogue, resulting in scaling issues. We can potentially amend this by inserting a \textit{Simulate} step after every certain number of \textit{Generate} steps, where each \textit{Simulate} step selects only the top $k$ candidates to continue generating upon, essentially implementing beam search on top of the dialogue trajectories. 

\subsection{Scalable episodic future thinking}
A key contribution of EGS is that it implements a novel scalable alternative to episodic future thinking. Human simulations of the future are limited by the linear stream of consciousness over time. However, simulated comparisons in EGS can be parallelized to achieve speeds much faster than human reasoning. 

In our experiments, each audience profile in each scenario is responsible for 1305 pairwise comparisons. This took between 2-4 hours and with one GPT-4 API key in September (10000 tokens per minute), with API queries costing around \$15. On average, this corresponds to one simulated pairwise comparison every 6-11 seconds, which may already surpass episodic future thinking in humans depending on the scenario. With recent rate limit increases and parallelization, this will only get faster, allowing for the use of EGS in delayed communication applications such as emails or public announcements. However, real-time applications such as improving conversational responses are still very limited. 

\subsection{The granularity of simulation} 
A key question that remains to be answered about audience simulation is whether there is a ``sweet spot'' for the level of detail a simulated audience should have. Currently, when we ask the LLM to generate stakeholder audiences, we do not specify the level of detail at which these stakeholders should be generated. 
However, we notice that many of the simulated comparisons (e.g., the \textit{Simulate} step in Table~\ref{tab:demo_table}) include a disclaimer about how the audience preference will depend on their personal details, which suggests that using a more detailed level granularity when generating audiences may improve the performance of the framework. At the same time, a higher level of detail or personalization could result in less accurate simulations due to the simulated situation being less common. We note that even with the current setup, the LLM does not enter the first-person perspective on some occasions, and adding more details might make this behavior more frequent. 

This effect is not just limited to the audiences, but also the details in the scenario itself. Various details of the real scenario can be included in the description fed into EGS. For instance, the plane crash example might be affected by whether there was a recent crash from the same company, or whether the weather conditions were bad that day, or even if people are generally more upset due to a global pandemic. Ultimately, there is a exciting trade-off between trying to accurately replicate the scenario and making the described scenario easier for the LLM to reason about, which we leave to future work.

\subsection{Application: Reasoning about the past}
Aside of optimizing communication in the present or preparing advice for the future, EGS can also be applied to reason about past communication events. In particular, we are able to perform counterfactual reasoning (CFR): Given a past scenario and its outcome, CFR concerns whether an alteration to the antecedent of the counterfactual affects the outcome~\citep{pearl2000causality}. While \citet{ma2023let} also use LLMs to do counterfactual reasoning, they focus on improving moral reasoning, whereas we propose the use counterfactuals to reason about causal effects in communication settings. 

Specifically, given a past communication setting, we can use the \textit{Explore} and \textit{Generate} components to create a diverse list of alternatives to the antecedent, and then \textit{Simulate} the outcomes when we replace the antecedent with each alternative. 
In this particular application, we can further improve the \textit{Simulate} step by including the communication utterance used and the actual outcome as a gold standard example in the context provided to the LLM. 
Then, using the simulated pairwise comparison results, we can make conclusions in the simulated space about which utterances the communication could have used to reach a better outcome, or if any underlying pieces of advice were responsible for a particular type of outcome. Though these causal effects may not be directly transferrable to the real world due to simulation inaccuracies, they provide testable hypotheses that can be directly implemented into real subject experiments.

\subsection{Application: Human studies and RLHF}
Aside of suggesting a communication utterance or advice, EGS can also be used directly for its simulated human preferences or feedback. In particular, we highlight two potential use cases in human subject studies and RLHF.

By viewing the instructions of human subject studies as ``communication'' with the goal of collecting high-quality data, EGS can be used to design and test a manifold of human user studies. 
Specifically, we can \textit{Explore} and \textit{Generate} different study design protocols, and then \textit{Simulate} participants to collect study data. Importantly, the simulated reactions of participants can be decoupled into (1) data collected for the study and (2) preferences on which study design is better, allowing for both design optimization and data collection at once. 
Many existing works show that LLMs are effective proxies of humans in research studies (see Section~\ref{sec:related_work}). However, these works stop at using LLMs to generate data/feedback, whereas EGS is also able to test and automatically select a best set of participant instructions as well. 

Unlike RLHF, which requires training explicit reward models~\citep{ouyang2022training}, EGS can generate audience members that function in some scenarios like reward models that rate generations. Furthermore, EGS is able to simulate audiences conditioned on different backgrounds and experiences, potentially allowing for better diversity and representation in the human feedback provided. 
Future work could present comparisons or collaborations between these two methods.

\subsection{LLMs as a representation of shared cultural experience}
By constructing a framework to help assist communication, we also make it possible to share information through the LLM's training data that may surpass individual, cultural, or geographical barriers. For example, a barista may not pay attention to a customer giving a coffee shop halfway across the globe a five star online review while describing their pleasant experience, but LLMs have the capability to take this information into account and synthesize a range of perspectives into the responses that they generate. Similarly, a person trying to solve a problem in their marriage might not have the habit of reading reddit forums on relationship advice, but LLMs might be able to take inspiration from these sources and use them to provide meaningful insights. In constructing this framework, we also hope to connect people with communication strategies that might not normally be available to them, not just helping them to improve their communication but also potentially helping them grow as communicators. 

\subsection{Broader impact and potential negative use}

Using scenarios covering the ten fundamental processes of interpersonal communication and internet user simulations in five diverse domains, we show that EGS allows for the improvement of a wide range of goal-oriented communication. EGS can potentially (1) improve communication with new audiences or people of unfamiliar backgrounds when mental simulations are uncertain, and (2) reduce misunderstandings by reasoning about message candidates from an audience's point of view. In particular, we support the use of EGS to allow ideas to be shared in ways that are more considerate and acceptable, especially between groups that are typically divided in their opinions.

We acknowledge that our framework, like most applications of LLMs, are dual-use and can also be used to optimize communications that may be detrimental to society. For instance, while EGS can be used simulate receivers to improve communication efficiency over emails, it can also be used in phishing emails to improve the chances of someone getting scammed. 
However, models are currently being fine-tuned and adjusted using RLHF to recognize and refuse to answer queries with malicious intentions~\citep{touvron2023llama, huang2023catastrophic}, and we can expect these efforts to positively impact the safety concerns of EGS as well. 

We also acknowledge that our constructed scenarios are about characters that fulfill common stereotypes, e.g., the white lie scenario being of a young man trying to compliment his female date. 
We encourage future work to analyze the performance of simulated audiences with more diverse backgrounds and relationships, especially since they are less represented in the training data which may affect their simulation accuracy. 
As a system that uses LLMs as a component, EGS may also adopt their inherent weaknesses including social biases that may seep into decision making. Thus, we recommend users to validate the outputs of EGS before putting them into use. 

Lastly, we would like to acknowledge how individuals in society may have unequal access to our method due to information barriers or financial costs (\$15 per simulated audience per scenario), and that this may increase effects of inequality. To partially alleviate this, we provide our code at \url{https://github.com/theryanl/EGS}. 

\section{Conclusion}

In this paper, we explored using LLMs to simulate audiences to help improve arbitrary goal-oriented communication. To do so, we introduced the EGS framework to \textit{Explore} the solution space, \textit{Generate} a large set of candidates, and \textit{Simulate} reactions of generated audiences to identify the best communication candidate and advice set. We found that EGS successfully generates and selects high-performing advice pairings and candidates, outperforming baselines such as GPT-4 with Chain-of-Thought. 
Furthermore, EGS has high agreement with human raters in a variety of scenarios, and extends well to broader online domains, offering a new and scalable method for simulating audience reactions to help us improve how we communicate with each other. 

Our results suggest that there may be other opportunities to draw on ideas from the cognitive science literature to create procedures that replace steps that might otherwise be internally performed by people with simulations generated by LLMs. By taking this approach, we can create systems that expand the capacities of LLMs and offer a novel form of support to humans seeking to communicate with and reason about other people.

\section*{Acknowledgements}
We thank the following colleagues for their great advice and thoughtful opinions: Mengzhou Xia, Ilia Sucholutsky, Yuhan Liu, Jiafei Duan, Sebastin Santy, Jieyu Zhang, and Joon Sung Park.

\bibliography{references}
\bibliographystyle{iclr2024_conference}

\appendix
\section{Appendix}

\subsection{Ten Fundamental Processes in Interpersonal Communication}
\label{app:ten_processes}
In this section, we provide more in-depth descriptions of the ten fundamental processes of interpersonal communication \citep{Berger2019Interpersonal}. These ten fundamental processes are seen as processes that underlie social interaction, and can be present regardless of the social context of the communication itself. 
\begin{itemize}
    \item Social Influence: When people's opinions or behavior are not in alignment, and they try to influence each other. 
    \item Social Support: To effectively support those experiencing distress. 
    \item Relationship Development: Separated into development (including initiation), maintenance, and deterioration (disengagement) of relationships. 
    \item Deception: Includes how performing deception may alter behavior, and the degree to which individuals are skilled at identifying deception. White lies fall under this category. 
    \item Bargaining and Negotiation: As exchanges involve risk, especially when partners are unsure that they can trust each other, communication processes have evolved to ensure fair exchanges.
    \item Conflict Management: Managing the negative consequences of conflict and how individuals confront each other. 
    \item Conversation Management: This suggests that individuals in a conversation implicitly understand that each will contribute to the conversation and advance its point.
    \item Impression Management: Individuals form impressions of each other that can influence both the courses of their interactions as well as future interactions and decisions. 
    \item Privacy Management: This involves the causes and consequences of self-disclosure, which is the tendency for individuals to reveal personal information to others. 
    \item Uncertainty Management: As individuals cannot be completely certain of their conversational partners' current emotional states, beliefs, attitude, and future actions, they engage in social interaction under conditions of uncertainty. 
\end{itemize}

\subsection{Scenario Choices and Details}
\label{app:scenarios}
In this section, we provide the descriptions for the eight scenarios we construct to evaluate our framework, followed by the fundamental processes they cover and their interesting qualities. The first four scenarios focus on more formal relationships, while the latter four are from more personal relationships. The scenarios also span various venues of communication, from speaking to written to digital. In constructing these scenarios, we hope to cover a wide range of potential situations that our framework could help assist people in. 

\paragraph{Plane Crash}
An airline spokesperson wants to minimize the negative public opinion of their company. Earlier today, a flight of the airline crashed due to unknown causes and there were 2 passengers that died. Right now, they are about to make the first public statement to the media at a live press conference. They only have 30 seconds of airtime to speak about the issue. 

The scenario focuses on the fundamental process of social support, while also covering impression management and social influence. A key objective in minimizing the negative public opinion is to provide support to the families and friends of the victims that are in experiencing distress. 
It also embodies a case where experimenting with the message is too risky for the agent and their company. Thus, within the large area of unexplored possibilities, there may be a better message to communicate available. Through the simulation framework we are able to ignore the risks and arrive at a reasonable outcome.

\paragraph{Product Launch}
A company CEO is announcing their new smartphone product, the jPhone, and they want to maximize its sales. The jPhone has the following features: \\
- Processing speeds equal to the top smartphones on the market\\
- Priced at \$700, which is 10-30\% cheaper than competitor smartphones on the market\\
- A unique, innovative design and user interface\\
Right now, they are about to give a quick 30-second presentation about the jPhone, broadcasted to major television channels. 

This scenario focuses on the fundamental process of social influence and impression management, but also covers the seller's facet in the bargaining and negotiation process, as one of the generated pieces of advice includes offering a flash sale for the first customers to purchase the jPhone.

\paragraph{Bargaining}
Jill wants to buy a round flower vase at a local artisan market. The marked price for the vase is 30 dollars. Jill would be willing to buy the vase at its original price, but she would also like to spend as little money as possible while still purchasing the vase. Right now, she is about to speak to the artisan selling the vase. 

This scenario focuses on the fundamental process of bargaining and negotiation, while also covering social influence and uncertainty management, as Jill does not know much the artisan is willing to lower their price. Together with the product launch scenario, we consider communication optimization from both seller and buyer perspectives.

\paragraph{Barista}
A barista at a coffee shop wants to maximize their tips received from customers. During their shift, a customer walks to the counter and says, "Hello, can I have a latte with whole milk, medium size?". The barista has a few seconds to respond. 

This scenario primarily covers the social influence fundamental process, while also including uncertainty management and conversation management. Customers arriving at a coffee shop may provide surface-level cues as to how willing they are to tip relative to their norm, but there is still uncertainty to what they would typically tip. Furthermore, this scenario embodies a case where the best conceptual knowledge may not be widely shared across geographical or cultural distances, but has a potential to improve the quality of life of both customers and baristas alike.

\paragraph{Sharing Secrets}
Mary is acquaintances with Carla through her job. She is currently chatting with Carla after work, and she wants to become closer friends with Carla. She thinks sharing a secret of hers might bring them closer together. The secrets that she can think of sharing, from least to most personal, are: \\
1. She used to hate pickles as a child but they have grown on her. \\
2. She has recently been feeling unconfident about her ability to succeed in her job. \\
3. She used to be impulsive when spending money and still has some debt. \\
Right now, she feels like it is a good point in the conversation to share a secret. She decides to try and do so. 

This scenario primarily focuses on the privacy management fundamental process, but also covers relationship development, conversational management, and impression management. This scenario differs from the others as there are a few options for the model to decide between, which aligns with the goal of being authentic when developing closer relationships. Furthermore, a selection paradigm is different from the original generation paradigm as the solution space is much more restricted, allowing for different qualities of the framework to be evaluated. 

\paragraph{Dating App}
A young man named Eric is in his early 20s and wants to find a girlfriend. He is currently an undergraduate student at a large state university, and likes to play tennis and hang out with friends in his free time. He likes animals and has a beagle dog named Scott. He would rate his looks and height at about average compared to those around him. He has decided to try his luck on a dating app. Right now, he is drafting his profile bio, and he is wondering what to write. 

This scenario primarily focuses on impression management and uncertainty management, with a smaller emphasis on the initiation stage of relationship development. On a dating app, who views your profile is unknown to you beforehand, and potential matches may vary widely in personality and what they prefer. Thus, the framework would need to take into account this uncertainty when optimizing for the best communication.

\paragraph{White Lie During Date}
A teenage guy is currently out on a date with a girl at the mall. He is attracted to the girl and wants to make the date as enjoyable for her as possible. They are currently in a clothing store and he is waiting while his date is trying out an outfit in the changing room. She comes out of the changing room looking very pleased while wearing an black denim jacket, strikes a pose, and asks, "what do you think!". Secretly, he thinks the jacket doesn't look good on her at all, but decides he should try to compliment it anyway. Right now, he is wondering what to say. 

This scenario primarily focuses on the deception fundamental process, while also covering relationship development and impression management processes. Deception is an important part of our lives, and sometimes we need to lie about our impressions in order to achieve our goals. We know that state-of-the-art LLMs can lie through hallucinations, but to our best knowledge lying intentionally has yet to be studied. 

\paragraph{Marriage Argument}
A woman is currently upset with her husband about the fact that he is consistently staying out with his friends until late at night instead of coming home to spend time with her. However, her husband is quite stubborn, and does not like to admit that he is in the wrong. She knows that her husband would be able to see reason eventually, but wants to minimize the time that it takes for him to see reason. Right now, she is about to approach him to talk about the issue. 

This scenario primarily focuses on the conflict management, while also including the maintenance aspect of the relationship development process and the social influence process. In delicate situations such as these, simulations can be very helpful in providing feedback and suggestions to the user.

\subsection{Audience Generation Prompts}
\label{app:stakeholders}
We query the LLM to generate the potential audiences and their corresponding weights and profiles as follows. First, we query the LLM for stakeholder audiences and weights: \\

\textbf{Prompt:} Please reason about and provide a list of up to $x$ stakeholders that are most relevant to the main character's goal, as well as a numerical value from 1 to 100 weighting the relative importance of each stakeholder. You do not need to reach the maximum amount of stakeholders, but instead include only stakeholders that would have an influence on the main character's goal. State your answer in the following format:\\
1. stakeholder name (stakeholder weight): justification\\
2. ...\\

Then, once we have the stakeholders, we generate their system and user prompts. \\

\textbf{Prompt:} Please write a description of the following role for the following scenario. It is important for the description to include a communication from the main character of the scenario. Here are a few examples for various stakeholders and scenarios:

[few shot examples]

Your task: 

Scenario: [provided]

Role: [provided]

Description: \\

\textbf{Prompt:} You are asking a question to a stakeholder to have them choose between the better of two scenarios. It is important that the question be tied directly to the goal that the main character of the story wants to achieve. Here are a few examples for various stakeholders and stories:

[few shot examples]

Your task: 

Story: [provided]

Stakeholder: [provided]

Question: \\

The above prompt generates a question to fill in the following template: \\

Scenario 1:
[scenario 1 candidate]

Scenario 2:
[scenario 2 candidate]

[insert question] Please think step by step and explain your reasoning. 
Put your response in the following format:
[reasoning text]

Answer: Scenario [1 or 2]

Then, these two are used as system prompt and user prompt to generate pairwise comparisons from the LLM. 

\subsection{Data Collection Procedure}
\label{app:experiment_procedure}
After providing an informed consent, participants received the following instructions: 
\begin{quote}
    ``Thank you for participating in our study! In this study, we will provide you with a series of scenarios, where in each scenario an individual will make a statement related to the scenario. Your task is to evaluate the quality of that statement with respect to the given scenario. You will have eleven response options, ranging from 0 (worst) to 10 (best). Choose the one you think is most appropriate. Note: no prior expertise is required to complete this task, just choose what you intuitively think is the right answer.'' 
\end{quote}
Participants then proceeded to the main experiment where they provided response-quality judgments on a 0-10 Likert scale for up to 20 randomly assigned responses. We detail the evaluation questions for each scenario below. 
\begin{itemize}
    \item Plane Crash: ``Based on the spokesperson's response, how do you think the company's public opinion would change?''. 
    
    \item Product Launch: ``If you were considering getting a new phone, how likely are you to buy a jPhone following the CEO's presentation?''. 
    
    \item Bargaining: ``Based on Jill's response, how likely would you be willing to negotiate the price of the vase with Jill?''. 
    
    \item Barista: ``Based on the barista's response, how likely are you to tip them?''. 
    
    \item Getting Close by Sharing Secrets: ``As Carla, how much closer would you feel you are with Mary after what she shared?''. 
    
    \item Dating App: ``Based on Eric's decision, how likely do you think his profile would attract a potential girlfriend?''. 
    
    \item White Lie During Date: ``As the girl, how do you think the guy's response would affect how much you are enjoying the date?''. 
    
    \item Marriage Argument: ``As the husband, how likely are you to admit that you are wrong immediately, instead of potentially defending yourself first or being dissatisfied at your wife's comments?''.
\end{itemize}

\begin{table}[]
\centering
\caption{Our human ratings have high inter-rater reliability (IRR) across all scenarios. For each scenario, we provide the number of candidate messages evaluated, the number of crowdworkers recruited, and IRR with 95\% confidence intervals. }
\vspace{10pt}
\label{tab:human_experiments}
\resizebox{0.69\textwidth}{!}{%
\begin{tabular}{@{}lccc@{}}
\toprule
Scenario                          & \# Candidates & \# Participants & IRR \\ \midrule
Plane Crash                       & 72                                                         & 76                       & $0.96 \pm 0.01$                                        \\
Product Launch                    & 72                                                         & 78                       & $0.75 \pm 0.08$                                             \\
Bargaining                        & 72                                                         & 78                       & $0.82 \pm 0.06$                                         \\
Barista                           & 102                                                         & 103                       & $0.90 \pm 0.03$                                            \\
Sharing Secrets  & 72                                                         & 79                       & $0.77 \pm 0.07$                                          \\
Dating App                        & 75                                                         & 80                       & $0.66 \pm 0.10$                                          \\
White Lie During Date             & 72                                                         & 80                       & $0.90 \pm 0.03$                                             \\
Marriage Argument                 & 72                                                         & 78                       & $0.77 \pm 0.08$                                          \\ \bottomrule
\end{tabular}%
}
\end{table}
Altogether, the number of participants, candidates, and the inter-rater reliability (IRR) with Spearman-Brown correlation~\citep{brown1910some} for each scenario is provided in Table~\ref{tab:human_experiments}. We observe a high IRR across all scenarios, pointing to the quality in the human data collected.

\subsection{Inter-rater reliability and agreement on unorthodox advice and other data subsets}
\label{app:irr}
\begin{table}[]
\centering
\caption{Inter-rater reliability (IRR) of candidates containing unorthodox advice is lower than any other subgroup. Baselines denotes IRR for GPT-4 zero-shot, GPT-4 Chain-of-Thought, and irrelevant and conceptual search space candidates. Normal Only denotes candidates without unorthodox advice in their advice sets. }
\vspace{10pt}
\label{tab:IRR}
\resizebox{0.67\textwidth}{!}{%
\begin{tabular}{@{}llcr@{}}
\toprule
Scenario                         & Baselines                & Normal Only              & Unorthodox               \\ \midrule
Plane Crash                      & 0.96 $\pm$ 0.01          & 0.71 $\pm$ 0.10          & 0.70 $\pm$ 0.13          \\
Product Launch                   & 0.78 $\pm$ 0.05          & 0.21 $\pm$ 0.26          & -0.40 $\pm$ 0.59         \\
Bargaining                       & 0.80 $\pm$ 0.04          & 0.18 $\pm$ 0.30          & -0.50 $\pm$ 0.64         \\
Barista                          & 0.90 $\pm$ 0.02          & 0.45 $\pm$ 0.18          & -0.79 $\pm$ 0.83         \\
Sharing Secrets & 0.78 $\pm$ 0.05          & 0.65 $\pm$ 0.12          & 0.55 $\pm$ 0.19          \\
Dating App                       & 0.71 $\pm$ 0.06          & 0.59 $\pm$ 0.15          & 0.37 $\pm$ 0.27          \\
White Lie During Date            & 0.87 $\pm$ 0.03          & 0.49 $\pm$ 0.18          & 0.44 $\pm$ 0.25          \\
Marriage Argument                & 0.82 $\pm$ 0.04          & 0.45 $\pm$ 0.19          & 0.62 $\pm$ 0.18          \\ \midrule
\textbf{Mean}                    & \textbf{0.83 $\pm$ 0.03} & \textbf{0.47 $\pm$ 0.07} & \textbf{0.12 $\pm$ 0.21} \\ \bottomrule
\end{tabular}%
}
\end{table}
We compute the IRR for baseline generations, candidates conditioned on only normal advice, and candidates with at least one unorthodox advice (unorthodox candidates) and find that unorthodox candidates have a much lower IRR (see Table~\ref{tab:IRR}). First, this suggests that unorthodox advice is harder to simulate and judge consistently even across human agents. Additionally, this implies that EGS using any LLM trained on human-generated data would likely also be more inconsistent during simulated judgements. 

With this insight, we analyze the agreement between both normal, one unorthodox, and both unorthodox candidate pairs using both the pairwise winner vs. loser and the percentage agreement metric (see Table~\ref{tab:pairwise_appendix}). We find that comparisons between normal advice to have higher agreement than comparisons between unorthodox advice in both metrics, but note that the differences are small. Intriguingly, we find that comparisons between normal candidates and unorthodox candidates have a much higher agreement in both metrics, indicating a high agreement in both absolute difference in ratings (winner vs. loser) and proportion of ratings (percentage agreement). 
\begin{table}[]
\centering
\caption{Pairwise winner vs. loser and percentage agreement analysis find agreement is highest in comparisons between normal and unorthodox candidates, then normal vs. normal candidates, and lowest in unorthodox vs. unorthodox candidates. There were 153, 216, and 66 both normal, one unorthodox, and both unorthodox comparisons respectively per scenario. }
\vspace{10pt}
\label{tab:pairwise_appendix}
\resizebox{\textwidth}{!}{%
\begin{tabular}{@{}lcccccc@{}}
\toprule
                                                          & \multicolumn{3}{c}{Winner vs. Loser}                                                                                                                                      & \multicolumn{3}{c}{Percentage Agreement}                                                                                                                                                  \\ \midrule
\begin{tabular}[c]{@{}l@{}}Compared\\ Advice\end{tabular} & \begin{tabular}[c]{@{}c@{}}Both \\ Normal\end{tabular} & \begin{tabular}[c]{@{}c@{}}One\\ Unorthodox\end{tabular} & \begin{tabular}[c]{@{}c@{}}Both \\ Unorthodox\end{tabular} & \begin{tabular}[c]{@{}c@{}}Both \\ Normal\end{tabular} & \begin{tabular}[c]{@{}c@{}}One\\ Unorthodox\end{tabular} & \begin{tabular}[c]{@{}c@{}}Both \\ Unorthodox\end{tabular} \\ \midrule
Plane Crash                                               & \hspace{15pt}0.38***                                                & \hspace{15pt}0.32***                                                  & \hspace{5pt}0.28*                                                      & 0.63                                                   & 0.63                                                     & 0.62                                                       \\
Product Launch                                            & 0.11                                                   & \hspace{15pt}0.55***                                                  & 0.13                                                       & 0.56                                                   & 0.81                                                     & 0.46                                                       \\
Bargaining                                                & \hspace{15pt}0.29***                                                & \hspace{12pt}-0.36***                                                 & \hspace{-3pt}-0.06                                                      & 0.70                                                   & 0.41                                                     & 0.53                                                       \\
Barista                                                   & \hspace{-3pt}-0.01                                                  & \hspace{15pt}2.33***                                                  & \hspace{-3pt}-0.04                                                      & 0.54                                                   & 0.75                                                     & 0.53                                                       \\
Sharing Secrets                                           & \hspace{15pt}0.56***                                                & \hspace{15pt}0.88***                                                  & \hspace{15pt}0.61***                                                    & 0.73                                                   & 0.82                                                     & 0.79                                                       \\
Dating App                                                & \hspace{-3pt}-0.15                                                  & \hspace{12pt}-0.26***                                                 & \hspace{-3pt}-0.15                                                      & 0.46                                                   & 0.37                                                     & 0.44                                                       \\
White Lie During Date                                     & 0.01                                                   & \hspace{2pt}-0.15*                                                   & \hspace{2pt}-0.27*                                                     & 0.47                                                   & 0.41                                                     & 0.37                                                       \\
Marriage Argument                                         & 0.13                                                   & \hspace{15pt}0.45***                                                  & \hspace{10pt}0.41**                                                     & 0.53                                                   & 0.73                                                     & 0.65                                                       \\ \midrule
\textbf{Mean}                                             & \textbf{0.16}                                          & \textbf{0.47}                                            & \textbf{0.11}                                              & \textbf{0.58}                                          & \textbf{0.62}                                            & \textbf{0.55}                                              \\ \bottomrule
\end{tabular}%
}
\end{table}

\subsection{Ablation on number of advice per candidate}
\label{multiple_advice}
\begin{table}[]
\centering
\caption{Having multiple advice per candidate does not consistently improve the mean candidate score, but often contributes to improving the best candidate. Errors shown with mean scores are standard errors of the mean, and *, **, *** denote significance at $p=0.05$, $0.01$, and $0.001$ respectively. }
\vspace{10pt}
\label{tab:ablation_num_advice}
\resizebox{0.94\textwidth}{!}{%
\begin{tabular}{@{}lllcc@{}}
\toprule
Scenario                         & \begin{tabular}[c]{@{}l@{}}Mean Score\\ Single Advice\end{tabular} & \begin{tabular}[c]{@{}l@{}}Mean Score\\ Multiple Advice\end{tabular} & \multicolumn{1}{l}{\begin{tabular}[c]{@{}l@{}}Highest Score\\ Single Advice\end{tabular}} & \multicolumn{1}{l}{\begin{tabular}[c]{@{}l@{}}Highest Score\\ Multiple Advice\end{tabular}} \\ \midrule
Plane Crash                      & \textbf{6.25 $\pm$ 0.20*}                                            & 5.88 $\pm$ 0.14                                                        & \textbf{7.95}                                                                               & 7.05                                                                                          \\
Product Launch                   & 5.71 $\pm$ 0.15                                                      & \textbf{6.24 $\pm$ 0.12**}                                             & 6.55                                                                                        & \textbf{7.05}                                                                                 \\
Bargaining                       & 5.77 $\pm$ 0.11                                                      & \textbf{6.06 $\pm$ 0.14}                                               & 6.30                                                                                        & \textbf{7.20}                                                                                 \\
Barista                          & \textbf{4.83 $\pm$ 0.53***}                                          & 3.57 $\pm$ 0.40                                                        & 6.20                                                                                        & \textbf{6.25}                                                                                 \\
Sharing Secrets & \textbf{5.38 $\pm$ 0.31}                                             & 5.34 $\pm$ 0.16                                                        & \textbf{7.30}                                                                               & 6.60                                                                                          \\
Dating App                       & \textbf{5.55 $\pm$ 0.17*}                                            & 5.20 $\pm$ 0.17                                                        & 6.40                                                                                        & \textbf{6.50}                                                                                 \\
White Lie During Date            & \textbf{6.80 $\pm$ 0.20}                                             & 6.72 $\pm$ 0.13                                                        & 7.95                                                                                        & \textbf{8.00}                                                                                 \\
Marriage Argument                & \textbf{6.32 $\pm$ 0.27}                                             & 6.09 $\pm$ 0.16                                                        & \textbf{7.80}                                                                               & 7.45                                                                                          \\ \bottomrule
\end{tabular}%
}
\end{table}
We investigate whether having multiple pieces of advice per candidate is beneficial for performance, and find that incorporating multiple pieces of advice per candidate 1) does not generally improve the mean performance, but 2) does improve the performance of the highest-performing candidate. In Table~\ref{tab:ablation_num_advice}, we provide the average scores and highest scores for candidates generated using 1 and 2 pieces of advice. We observe that, even though single advice performs better on average than multiple advice (but not generally significantly so), multiple advice is responsible for the human-rated top candidate in five of the eight scenarios. This demonstrates the benefit of conditioning candidates on combinatorial sets of advice in the \textit{Generate} step.

\subsection{Simulated audience judgements contain no evidence of positional bias}
\label{app:positional_bias}
In past work~\citep{wang2023large}, LLMs are shown to have a tendency of preferring some options over others simply due to the order they appear in the list of options. We conduct an analysis on whether this behavior can be detected in pairwise comparisons done by simulated audiences. Across the eight scenarios, we assign ``prefer scenario 1'' with value 1, ``prefer scenario 2'' with value 0, and ``tie'' with value 0.5, similar to Equation~\ref{eq:pairwise}. We take the average over all pairwise comparisons and find a mean of $0.5011$ and a standard error of the mean of $0.0037$. This leads us to conclude that there is no evidence of a positional bias in the pairwise comparisons of simulated audiences.

\subsection{Agreement becomes more pronounced for easier cases}
\label{app:large_differences}
\begin{table}[]
\centering
\caption{Agreement between GPT-4 pairwise ratings and human scores (compared pairwise). We consider the set of all pairwise comparisons, as well as subsets where the absolute difference of the human ratings, $d$, differ by at least 0.5, 1, and 1.5 respectively. $n$ denotes the number of samples in the corresponding subsets. 
}
\vspace{10pt}
\label{tab:agreement_large_differences}
\resizebox{\textwidth}{!}{%
\begin{tabular}{@{}lcc|cc|cc|cc@{}}
\toprule
Scenario                        & \begin{tabular}[c]{@{}c@{}}Agreement\end{tabular} & $n$ & $d \geq$ 0.5 & $n$ & $d \geq 1$ & $n$ & $d \geq$ 1.5 & $n$ \\ \midrule
Plane Crash                     & 0.63                                                          & 435 & 0.72                                                            & 259 & 0.80                                                          & 122 & 0.83                                                            & 46  \\
Product Launch                  & 0.67                                                          & 435 & 0.74                                                            & 237 & 0.83                                                          & 92  & 0.96                                                            & 28  \\
Bargaining                      & 0.53                                                          & 435 & 0.43                                                            & 243 & 0.41                                                          & 86  & 0.33                                                            & 24  \\
Bargaining (-unorthodox advice) & 0.69                                                          & 153 & 0.79                                                            & 72  & 0.88                                                          & 21  & 1.00                                                            & 3   \\
Barista                         & 0.64                                                          & 435 & 0.71                                                            & 330 & 0.77                                                          & 269 & 0.81                                                            & 239 \\ 
Sharing Secrets                 & 0.78                                                          & 435 & 0.84                                                            & 312 & 0.91                                                          & 194 & 0.94                                                            & 100 \\
Dating App                      & 0.41                                                          & 435 & 0.37                                                            & 282 & 0.31                                                          & 144 & 0.37                                                            & 52  \\
White Lie During Date           & 0.43                                                          & 435 & 0.41                                                            & 265 & 0.40                                                          & 124 & 0.58                                                            & 33  \\
Marriage Argument               & 0.65                                                          & 435 & 0.64                                                            & 312 & 0.71                                                          & 181 & 0.74                                                            & 89  \\ \bottomrule
\end{tabular}%
}
\end{table}
We consider easier pairwise comparison cases where the difference in mean ratings between the two candidates are greater than certain thresholds, and find that agreement between humans and LLM-simulated audiences becomes more pronounced for both high and low agreement cases (see Table~\ref{tab:agreement_large_differences}). 
We calculate agreement using the percentage agreement metric proposed in Section~\ref{sec:percentage_agreement}. 
For scenarios starting with agreement \textgreater~0.6 (plane crash, product launch, bargaining (-unorthodox), barista, sharing secrets, marriage argument), as we increase the threshold agreement consistently increases as well. In particular, five of these reach $0.8$ agreement with moderate sample sizes. 
For the three scenarios that begin with agreement \textless 0.6, agreement generally continues to decrease as the threshold increases, indicating a fundamental misalignment between human and simulated audience judgments.

\subsection{SHP Experiments}
\label{app:shp}

Following the steps taken by the original authors of the SHP dataset\footnote{\url{https://huggingface.co/datasets/stanfordnlp/SHP}}, we first filter the dataset such that all pairwise comments have a ratio of at least $3$ in the number of upvotes. 
This aims to reduce the amount of noise in the dataset and ensure that one comment is strongly preferred over the other.
Furthermore, we randomly assign the order of the comments such that the chance of any comment being either the first or the second comment is 50\%. This reduces possible positional biases from the model.
For generation, we use hyperparameters \texttt{top\_p=0.9} and \texttt{temperature=0.1}.

\begin{table}[ht]
    \centering
    \small
    \caption{Prompt used in the CoT setting.}
    \vspace{10pt}
    \begin{tabular}{l}
    \toprule
    {\color{blue}User: }\\
    Comment 1:\\
    {\color{brown}\textless Comment 1\textgreater} \\
    Comment 2:\\ 
    {\color{brown}\textless Comment 2\textgreater} \\
    Post: \\
    {\color{brown}\textless Post\textgreater} \\
    Given the post, choose the comment that you are more likely to upvote.\\
    Please think step by step and explain your reasoning. Stop after you output the final answer.\\
    Put your response in the following format:\\
    Reason: [reasoning text]\\
    Answer: Comment [1 or 2]\\
    End\\
    \bottomrule
    \end{tabular}
    \label{tab:shp-cot-prompt}
\end{table}

\begin{table}[ht]
    \centering
    \small
    \caption{Prompt used in the Redditor Simulation -- Default setting for the askculinary domain.}
    \vspace{10pt}
    \begin{tabular}{l}
    \toprule
    {\color{blue}System: }\\
    You are interested in all culinary-related things. 
    \\You are currently browsing a reddit culinary forum, and you are looking for interesting content to read. 
    \\You click on a post and you are reading through the comments.\\
    \midrule 
    {\color{blue}User: }\\
    Comment 1:\\
    {\color{brown}\textless Comment 1\textgreater} \\
    Comment 2:\\ 
    {\color{brown}\textless Comment 2\textgreater} \\
    Post: \\
    {\color{brown}\textless Post\textgreater} \\
    Given the post, choose the comment that you are more likely to upvote.\\
    Please think step by step and explain your reasoning. Stop after you output the final answer.\\
    Put your response in the following format:\\
    Reason: [reasoning text]\\
    Answer: Comment [1 or 2]\\
    End\\
    \bottomrule
    \end{tabular}
    \label{tab:shp-redditor-sim-prompt}
\end{table}

\begin{table}[ht]
    \centering
    \small
    \caption{Prompt used in the Redditor Simulation -- Funny setting for the askculinary domain.}
    \vspace{10pt}
    \begin{tabular}{l}
    \toprule
    {\color{blue}System: }\\
    You are interested in all culinary-related things. 
    \\You are currently browsing a reddit culinary forum, and you are looking for interesting and entertaining \\content to read. 
    \\You are more likely to upvote comments that are funny and that can make you laugh. 
    \\You click on a post and you are reading through the comments.\\
    \midrule 
    {\color{blue}User: }\\
    Comment 1:\\
    {\color{brown}\textless Comment 1\textgreater} \\
    Comment 2:\\ 
    {\color{brown}\textless Comment 2\textgreater} \\
    Post: \\
    {\color{brown}\textless Post\textgreater} \\
    Given the post, choose the comment that you are more likely to upvote.\\
    Please think step by step and explain your reasoning. Stop after you output the final answer.\\
    Put your response in the following format:\\
    Reason: [reasoning text]\\
    Answer: Comment [1 or 2]\\
    End\\
    \bottomrule
    \end{tabular}
    \label{tab:shp-redditor-sim-prompt-funny}
\end{table}

\begin{table}[ht]
    \centering
    \small
    \caption{Example of the output from the three different settings in the askculinary domain.}
    \vspace{10pt}
    \begin{tabular}{p{0.95\linewidth}}
    \toprule
    \textbf{Post}: \\
    When setting up a breading station, is it better to season the flour or panko? For example when you're breading proteins such as chicken or pork, do you season the flour or the panko? I've always just seasoned the flour but was curious to hear what other people do. \\\\
    \textbf{Comment 1 (42 upvotes)}:\\
    100\% read this as ``breeding station''.  Looked at subs name.  Asked myself what the fuck was wrong with people.  Reread the title.  Whats wrong with people is that some of us can't read.  As to your question. I do both, though they are usually lightly seasoned.\\
    \textbf{Comment 2 (9 upvotes)}:\\
    I season all steps.\\\\
    \midrule
    \textbf{Chain-of-Thought} Output\\
    \midrule
    Reason: Comment 2 is more relevant and directly answers the question posed in the post. It provides a clear opinion on the topic of seasoning in a breading station. Comment 1, while humorous, doesn't provide a clear stance on the question and spends more time discussing a reading error. 
    \\Answer: Comment 2\\
    \midrule
    \textbf{Redditor Simulation -- Default} Output\\
    \midrule
    Reason: Comment 1 provides a humorous anecdote before answering the question, which can be entertaining but doesn't directly address the post's question until the end. Comment 2, on the other hand, is straightforward and directly answers the question, providing a clear stance on the topic. As someone interested in culinary-related things, I would appreciate the direct and relevant response more.\\ 
    Answer: Comment 2\\
    \midrule
    \textbf{Redditor Simulation -- Funny} Output\\
    \midrule
    Reason: Comment 1 is more likely to be upvoted because it contains a humorous anecdote about misreading the post title, which adds an entertaining element to the discussion. While both comments provide an answer to the question, Comment 1 does so in a way that is more engaging and likely to elicit a laugh.\\
    Answer: Comment 1\\
    \bottomrule
    \end{tabular}
    \label{tab:shp-culinary-example}
\end{table}

\begin{table}[ht]
    \centering
    \small
    \caption{Example of the output from the three different settings in the asksocialscience domain.}
    \vspace{10pt}
    \begin{tabular}{p{0.95\linewidth}}
    \toprule
    \textbf{Post}: \\
    Why is the US so dominating in entertainment and culture? I'm from Norway, not the largest country in the world, and I was thinking about where my entertainment comes from. Speaking a small language, I'm accustomed to speaking or listening to another language when I interact online or watch/listen to other entertainment mediums, and I'd guess that most non-english speaking people would do the same.   But it seems like most of my entertainment comes from the US, something that I find weird when I think about the size of the country. Europe has more than double the population, and by that logic why isn't 2/3 of my entertainment from the continent? In global terms, the US is roughly 5\% of the world population, yet it's dramatically overrepresented in most media...\\\\
    \textbf{Comment 1 (5 upvotes)}:\\
    I'm from Russia and puzzled about it as well. Speaking about TV, for example \u2014 I never watch television here, because it's 99\% shit, but regurarly download latest episodes of Breaking Bad, Homeland and other great american shows... What I don't understand, personally, is how they ended up making stuff that's so much better than what we have here.\\\\
    \textbf{Comment 2 (16 upvotes)}:\\
    First: The Division of Labor is limited by the extent of the market. English language extends to a much larger number of wealthy people than any other language, which means there can be more differentiation within the English language market than other language markets... Second: Agglomeration effects and economies of scale*. Certain types of art thrive by being done at a place where other types of that art are being done. Obviously, if I want to make a movie the cheapest place to produce it will be Hollywood (filming it is a different question), because there are people and equipment there who specialize in making movies... Economies of Scale indicate the range of production where average cost is decreasing...\\
    \midrule
    \textbf{Chain-of-Thought} Output\\
    \midrule
    Reason: Comment 2 provides a detailed and well-thought-out explanation to the question posed in the post. It uses economic principles to explain why the US dominates in entertainment and culture, and gives examples to illustrate these principles. It also addresses the question from different angles, including the influence of language and the role of specific locations in the production of art. On the other hand, Comment 1 is more of a personal opinion and doesn't provide a comprehensive answer to the question.\\Answer: Comment 2\\
    \midrule
    \textbf{Redditor Simulation -- Default} Output\\
    \midrule
    Reason: Comment 2 provides a detailed and well-thought-out response to the original post. It offers a comprehensive explanation of why the US dominates in entertainment and culture, using concepts such as the division of labor, agglomeration effects, and economies of scale. It also provides examples to illustrate these concepts, making the explanation more understandable. On the other hand, Comment 1 is more of a personal anecdote and doesn't really answer the question posed by the original post. \\Answer: Comment 2 \\
    \midrule
    \textbf{Redditor Simulation -- Funny} Output\\
    \midrule
    Reason: While both comments provide insightful perspectives on the question posed by the post, Comment 2 offers a more detailed and comprehensive explanation. However, considering my preference for humor and entertainment, Comment 1 is more likely to get my upvote. It is more relatable and has a touch of humor in the way the commenter compares the quality of TV shows in their home country to American shows. \\Answer: Comment 1\\
    \bottomrule
    \end{tabular}
    \label{tab:shp-socialscience-example}
\end{table}

\subsection{Aggregation of Stakeholder Opinions}
\label{app:aggregation}

In this section, we conduct a preliminary investigation into alternative aggregation methods of stakeholder opinions. Our scenarios can include multiple LLM-generated stakeholders, and weights assigned to each by the LLM. However, a weighted sum is not necessarily the best approach for aggregating stakeholder opinions together. For instance, if all stakeholders are able to severely affect the communicator's goal, then they might opt for a strategy to maximize the lowest score across all stakeholders. Alternatively, if you only need one stakeholder's approval to achieve your goal, a viable strategy might be to maximize the highest score across all stakeholders. 

In the cognitive science literature, \citet{gates2020how} show that when humans are trying to be helpful to multiple people at once, their behavior is best described by the \textit{maximin} metric, describing the desire to maximize the happiness of the worst-off person, while also consistent with maximizing group utility (\textit{maxsum}) and equality (\textit{inequality aversion}) to a lesser extent. 

In our three scenarios with multiple stakeholders -- product launch, plane crash, and dating app, we find that these metrics agree with each other in all but two cases on the best advice and one case on the best candidate. Notably, the \textit{maximin} metric performs worse on the plane crash scenario, selecting the 17th best candidate (out of 30) instead of the 1st, and the 3rd best advice set instead of the 2nd. In the future, more investigations can be done to determine which of the aggregation metrics make the most sense on top of LLM comparisons. 

\end{document}